\pdfoutput=1
\PassOptionsToPackage{table}{xcolor}
\documentclass[11pt]{article}

\usepackage[final]{acl}

\usepackage{times}
\usepackage{latexsym}

\usepackage[T1]{fontenc}
\usepackage[utf8]{inputenc}

\usepackage{microtype}

\usepackage{inconsolata}

\usepackage[table]{xcolor}
\usepackage{graphicx}

\newcommand{\loads}[1]{\textsc{LOADS}}

\usepackage{booktabs}
\usepackage{multirow}
\usepackage{rotating}
\usepackage{subcaption}
\usepackage{amsmath,amssymb}
\usepackage{enumitem}

\title{Label Set Optimization via Activation Distribution Kurtosis for Zero-Shot Classification with Generative Models}

\author{Yue Li, Zhixue Zhao \and Carolina Scarton \\
        Department of Computer Science, University of Sheffield, UK\\
        \texttt{\{yli381,zhixue.zhao,c.scarton\}@sheffield.ac.uk}}

\begin{document}
\maketitle
\begin{abstract}
In-context learning (ICL) performance is highly sensitive to prompt design, yet the impact of class label options (e.g. lexicon or order) in zero-shot classification remains underexplored. This study proposes \loads{} (Label set Optimization via Activation Distribution kurtosiS), a post-hoc method for selecting optimal label sets in zero-shot ICL with large language models (LLMs).  
LOADS is built upon the observations in our empirical analysis, the first to systematically examine how label option design (i.e., lexical choice, order, and elaboration) impacts classification performance. This analysis shows that the lexical choice of the labels in the prompt (such as \textit{agree} vs. \textit{support} in stance classification) plays an important role in both model performance and model's sensitivity to the label order. A further investigation demonstrates that optimal label words tend to activate fewer outlier neurons in LLMs' feed-forward networks. \loads{} then leverages kurtosis to measure the neuron activation distribution for label selection, requiring only a single forward pass without gradient propagation or labelled data. The \loads{}-selected label words consistently demonstrate effectiveness for zero-shot ICL across classification tasks, datasets, models and languages, achieving maximum performance gain from 0.54 to 0.76 compared to the conventional approach of using original dataset label words.
\end{abstract}

\section{Introduction}

Generative large language models (LLMs) are increasingly used for classification tasks via zero-shot in-context learning (ICL), where models are prompted to select an option from a pre-defined set of labels~\cite{wang-etal-2022-super,antypas-etal-2023-supertweeteval,gonen-etal-2023-demystifying,mu-etal-2024-navigating}. While some classification tasks employ a relatively fixed set of lexicons to represent class labels, such as sentiment analysis (\textit{positive} and \textit{negative}) and textual entailment (\textit{entailment} and \textit{contradiction}), other tasks may present more ambiguous choices in lexical selection. Stance classification, for instance, uses diverse pairs of antonyms to represent positive and negative stances across different datasets, e.g. \textit{agree}-\textit{disagree} vs. \textit{favor}-\textit{against}. As a result, when crafting prompts for these classification tasks, practitioners face decisions regarding label options in the prompt, such as lexical selection and ordering.

\begin{figure*}[ht!]
 \centering
 \includegraphics[width=.8\textwidth]{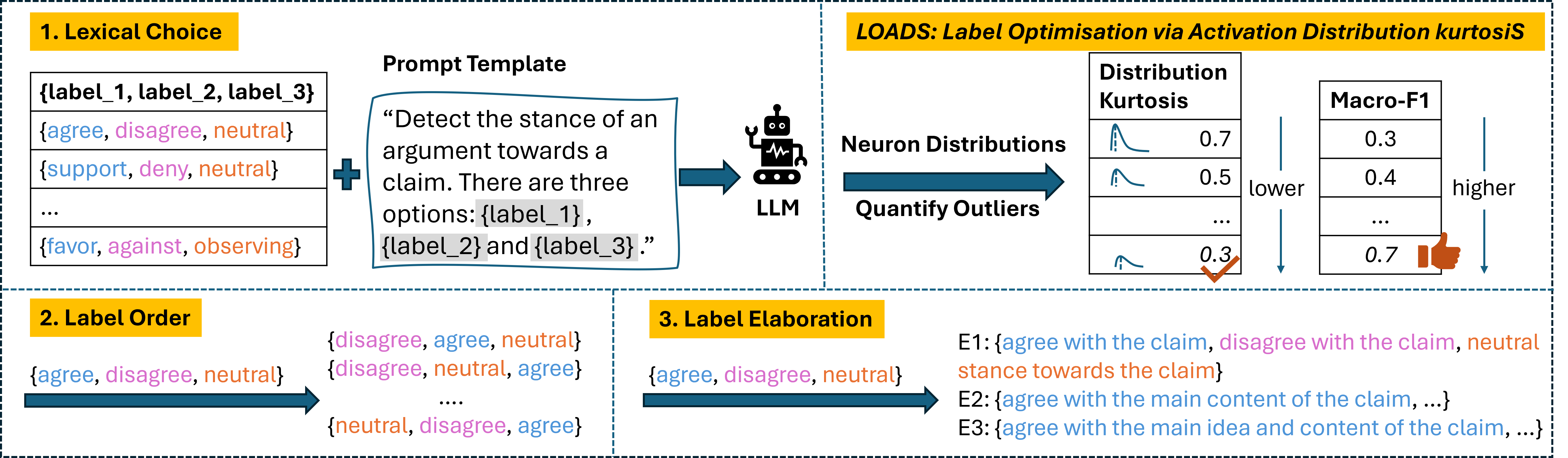}
 \caption{Illustration of the three aspects (i.e., lexical choice, label order and label elaboration) for designing the label option in the prompt in zero-shot ICL for classification, and our \loads{} to post-hoc select the optimal label set (top half figure).}
 \label{fig:overview}
\end{figure*}

Despite studies suggesting the sensitivity of ICL to prompt design~\cite{lu-etal-2022-fantastically,yoo-etal-2022-ground,10.5555/3600270.3602070,mao-etal-2024-prompt,liu-etal-2024-lost,zhang-etal-2022-active,liu-etal-2022-makes,peng-etal-2024-revisiting,gonen-etal-2023-demystifying,mu-etal-2024-navigating}, this subtle yet critical consideration of label options in prompt for zero-shot ICL has received limited attention. To fill in this research gap, we explore the impact of three types of label variants (i.e., lexical choice, label order and elaborations) in zero-shot ICL with both encoder-decoder and decoder-only LLMs. We mainly ground our research on stance classification, a task where label adaptation is a known problem due to various label inventories in different studies~\cite{hardalov-etal-2021-cross}. We demonstrate that the lexical choice of the label options significantly impacts model performance. The model's sensitivity to the label order also depends on the lexical choice, while
% and when the lexical choice of labels is suboptimal, the order of the labels also affects performance. 
elaborating on task-related information (e.g. \textit{agree with the claim} elaborating \textit{agree}) has minimum effect.

% one sentence to summarize findings, xx matter xx doesnot matter.
% add some findings and observations here

%Further analysis of the model internal state is also conducted to gain insights on optimal label selection. 
Inspired by recent studies on neuron analysis~\cite{kuzmin2023pruning,voita-etal-2024-neurons,stolfo2024confidence}, we further investigate the neurons in the feed-forward network (FFN) in the decoder of the LLMs. We empirically show that prompts with optimal label sets activate fewer outlier neurons. Consequently, we propose a new method, \textbf{L}abel set \textbf{O}ptimization via \textbf{A}ctivation \textbf{D}istribution kurtosi\textbf{S} (\loads{}), to select optimal label sets for a given classification dataset in zero-shot ICL. \loads{} could stably and effectively work with only 100 unlabelled samples of the validation dataset, also demonstrating transferability across datasets and languages. Our contributions are summarized as follows: 

\begin{itemize}[leftmargin=*, noitemsep]

    \item The first benchmark on \textbf{how variants of label options in prompts affect zero-shot ICL models' performance} for classification tasks. We provide useful recommendations on label designing to practitioners working on zero-shot classification with LLMs. 
    \item The empirical demonstration that \textbf{zero-shot ICL performance negatively correlates with the number of outlier neurons in FFN} when varying the lexical choices for label options. The correlation holds true across diverse English stance classification datasets and topic classification datasets with different models.
    \item A \textbf{novel and efficient post-hoc method (\loads{}) for label selection in zero-shot ICL}. Compared with common strategies in practice, our approach demonstrates statistically significant performance improvements across model types, model sizes and languages with only 100 unlabeled data samples. Our analysis also suggests that the \loads{}-selected label set is potentially transferable across similar datasets for a specific LLM, further alleviating the cost to collect samples for the target new dataset.

\end{itemize}

We present our experimental setups, results and discussions on the impact of label options in zero-shot ICL in Section \ref{sec:label setup}. Then, we describe our neuron analysis of the lexical choice in label options in Section \ref{sec:neuron} and our proposed method \loads{} for selecting optimal label sets in Section \ref{sec:load}.

\section{Related Work}

\paragraph{ICL Performance} \label{sec:related work}

Few-shot ICL mainly focuses on cases where LLMs are directly prompted with $N$ demonstration examples. The studies highlight the substantial impact of example ordering \cite{lu-etal-2022-fantastically}, formatting \cite{yoo-etal-2022-ground,10.5555/3600270.3602070,mao-etal-2024-prompt,liu-etal-2024-lost}, and examples selection \cite{zhang-etal-2022-active,liu-etal-2022-makes,peng-etal-2024-revisiting}. %Progress has also been made in 
The parallel lines of work focus on improving few-shot ICL via %different means, including 
the optimal selection or arrangement of examples~\cite{liu-etal-2022-makes,rubin-etal-2022-learning,lu-etal-2022-fantastically,zhang-etal-2022-active,liu2024let,xu-etal-2024-context}, re-weighting examples~\cite{yang-etal-2023-demonstration}, automatic reformat or generation of demonstration representations~\cite{kim2022self,liu2023context}, and introduction of intermediate reasoning steps~\cite{wei2022chain,zhang2023automatic}. However, the impact of lexical choices for label names in classification received little attention, with the only closely related work %in this line of research s
suggesting that LLMs are likely to confuse classes which share similar key vectors in the attention modules \cite{wang-etal-2023-label}.

For zero-shot ICL, Mu et al., \shortcite{mu-etal-2024-navigating} demonstrate the effect of using synonyms for class options, but they neither consider the order of the label options nor propose an effective strategy to choose the label names. Gonen et al., \shortcite{gonen-etal-2023-demystifying} empirically show that perplexity could be an effective indicator for prompt selection, but %their process 
they do not account for class options. Notably, behavioral differences between few-shot and zero-shot ICL have been frequently observed, suggesting that findings from few-shot ICL do not necessarily hold in the zero-shot context~\cite{lin2024dual}. 
% https://aclanthology.org/2024.findings-naacl.258.pdf prompt position

\paragraph{Prompt-Tuning and Verbalizer} Prompt-tuning aims to automatically find or generate an optimal discrete prompt (e.g., through gradient-based search \cite{shin-etal-2020-autoprompt,shi-etal-2023-toward} or fine-tuning \cite{gao-etal-2021-making,le-scao-rush-2021-many,deng-etal-2022-rlprompt}) or by training continuous prompt \cite{lester-etal-2021-power,liu-etal-2022-p}. 
%Verbalizer is commonly followed after prompt-tuning or pattern exploit training \cite{gao-etal-2021-making} for the language models pre-trained with masked language modelling. It 
Verbalizer can be taken as a mapping function that links discrete class labels 
%(e.g., ``positive," ``negative") 
to corresponding tokens or phrases in a model's vocabulary. %aims to map output predictions to labels for classification tasks, 
A range of methods developed to build the verbalizer, including manually created verbalizer \cite{schick-schutze-2021-exploiting}, search-based verbalizer that identifies label words automatically from the dataset \cite{gao-etal-2021-making,shin-etal-2020-autoprompt}, and soft verbalizers that uses continuous embeddings obtained through fine-tuning \cite{hambardzumyan-etal-2021-warp,cui-etal-2022-prototypical}. Prompt-tuning often does not focus on label set selection for zero-shot ICL, and the verbalizer introduces additional components to the decoding of the generative models, distinguishing it fundamentally from our work.

\section{Prompting with Varied Label Options for Zero-shot Classification} \label{sec:label setup}

In zero-shot ICL for classification, a common approach is to provide a set of class label options in the prompt to instruct the LLMs to choose one of the options as the classification prediction. Although the label option is a subtle component in the prompt, we are interested in whether it has a significant impact on model performance.

Specifically, we explore three types of variants around label options in the prompt: (1) \textit{lexical choice}; (2) \textit{label order}; and (3) \textit{label elaboration}. To accurately measure the impact of these factors, we only manipulate the label options within the same prompt template (Section \ref{sec:prompt_templte}). We show examples of the three variants in Figure \ref{fig:overview}.

\subsection{Methodology}
\paragraph{Lexical Choice} \label{sec:Lexical Choice}

We use single-word synonyms to represent class labels (e.g. \textit{support} and \textit{agree}), forming various label sets. For each dataset, we compare the zero-shot ICL performance when LLMs are prompted to select from different label sets, as illustrated in Figure \ref{fig:overview}. For this purpose, we design a pipeline to create a pool of label sets: 
\begin{enumerate}[leftmargin=*, noitemsep]
    \item \textit{Collect a seed set of label names.} We obtain this set by collecting the label names in the datasets we experiment with.
    \item \textit{Expand label sets with WordNet and LLMs.} We use WordNet \cite{fellbaum1998wordnet} as a reliable source and Claude\footnote{\url{https://claude.ai/new}} as a supplementary source to obtain synonyms for label names in the seed set. For pairs of label names with semantically opposite meanings (such as ``agree" and ``disagree"), we also consider antonyms to avoid potential ambiguity and present clear contrast for the predicted models.
    \item \textit{Manual selection.} We manually filter out semantically unrelated or inappropriate label sets generated by Claude to mitigate the impact of noisy label names.
\end{enumerate}

The label names are arranged in the sequence presented in their original study (see Table \ref{tab:datasets}). We refer to this arrangement as the \textit{default order}.

\paragraph{Label Order} \label{sec:label order}

We consider every possible order of the single-word labels in the prompt and compare the model performance against that obtained with the default order. For binary datasets, there is only one alternative arrangement besides the default order, while $N$-way multi-class classification would yield $N!-1$ alternative orders. 

\paragraph{Label Elaboration} \label{sec:label elaboration}

We investigate whether transforming single-word labels (e.g., ``agree") into more detailed phrases (e.g., ``agree with the claim") has an impact on model performance. On the one hand, elaborating on task details with the label may provide the model with a stronger alignment signal between the label and the task, emphasizing what the label is referring to. On the other hand, it also increases the label length and may introduce noise in the prompt~\cite{liu-etal-2024-lost}. Therefore, we design three levels of elaborations (shorted for \textit{E1}, \textit{E2} and \textit{E3}) by progressively adding more task-related (and potentially redundant) information to the single-word label names, as shown in Figure \ref{fig:overview}.

\subsection{Experimental Setups} \label{sec:exp setup}

\paragraph{Datasets} We focus on the stance classification task due to its rich label inventories across various readily available datasets. Stance classification aims to identify the type of an expressed opinion (e.g., ``agree" or ``disagree") in a given piece of text towards a particular topic, claim, or entity. We consider four binary (\textit{scd} \cite{hasan-ng-2013-stance},
\textit{perspectrum} \cite{chen-etal-2019-seeing},
\textit{snopes} \cite{hanselowski-etal-2019-richly} and
\textit{ibmcs} \cite{bar-haim-etal-2017-stance}) and five multi-class datasets (\textit{vast} \cite{allaway-mckeown-2020-zero},
\textit{emergent} \cite{ferreira-vlachos-2016-emergent},
\textit{semeval} \cite{mohammad-etal-2016-semeval},
\textit{rumoureval} \cite{gorrell-etal-2019-semeval} and
\textit{arc} \cite{hanselowski-etal-2018-retrospective}) from existing English stance classification benchmarks \cite{schiller2021stance,hardalov-etal-2021-cross,chen-etal-2023-close}, as shown in Table \ref{tab:datasets}. The nine datasets cover different domains, such as social media posts, news articles and online debates forums. We also experiment with topic classification in Section \ref{sec:neuron} and \ref{sec:load} to demonstrate the generalizability of our findings to other NLP tasks.
% The target of the stance varies from non-phrase topics (e.g., ``immigration" or ``school uniform") to phrases or statements (e.g., news headlines, claims or rumourous tweets). 
% Details can be found in the Appendix.
% Table \ref{tab:datasets} summarises these nine datasets. 

\begin{table}[t!]
\centering
\scalebox{0.55}{
\begin{tabular}{l|ll}
\hline
\textbf{Dataset Name} & \textbf{Original Label Words} & \textbf{Optimal Label Words}\\
\hline
scd & for, against
& pro, con\\
perspectrum  & support, undermine
& validate, refute\\
snopes  & agree, refute
& affirm, refute\\
ibmcs & pro, con & endorse, deny\\
\hline
vast & pro, con, neutral
& confirm, dispute, neither\\
emergent & for, against, observing
& endorse, reject, neutral\\
semeval & favour, against, neither
& accept, reject, neutral\\
rumoureval & support, deny, query, comment
& confirm, reject, question, neutral\\
arc & agree, disagree, discuss, unrelated
& affirm, refute, discuss, unrelated\\
\hline
\end{tabular}
}
\caption{Lists of the English stance classification datasets, their labels in \textit{original} dataset, and the \textit{optimal labels} with the highest zero-shot ICL performance on Flan-T5-xl  as an example to justify our motivation on \loads{}.}
\label{tab:datasets}
\end{table}

\paragraph{Models}\label{sec:prompt_templte}

We cover both encoder-decoder and decoder-only LLMs, and experiment with the prevalent open-sourced Flan-T5 \cite{chung2024scaling}, Llama 3 and Llama 3.1 \cite{dubey2024llama} model families as representatives for these two types of LLMs. We choose their moderate-sized instruction-tuned versions, Flan-T5-xl (3b), Llama -3-Instruct (8b) and Llama-3.1-Instruct (8b), as our primary models for investigation due to our hardware resources constraints and their decent zero-shot ICL performance~\cite{aiyappa2024benchmarking,chung2024scaling,dubey2024llama}. We conduct experiments with \textit{Gemma-2-it (9b)} \footnote{https://huggingface.co/google/gemma-2-9b-it} and \textit{Flan-T5-xxl} (13b) to further show the generalizability of \loads{} in Section \ref{sec:load}.

\begin{table*}[ht!]
\centering
\scalebox{0.6}{
\begin{tabular}{l|l|cccc|ccccc}
\hline
& \textbf{$F_{score}$ }& \textbf{perspectrum} & \textbf{ibmcs} & \textbf{snopes} & \textbf{scd} & \textbf{emergent} & \textbf{semeval} & \textbf{rumoureval} & \textbf{arc} & \textbf{vast} \\
\hline
\multirow{4}{4em}{Llama 3 (8b)} & \textit{max} & 0.869 & \cellcolor[HTML]{ADD8E6}0.834 & \cellcolor[HTML]{ADD8E6}0.770 & 0.759 & \cellcolor[HTML]{1E90FF}0.740 & 0.688 & \cellcolor[HTML]{1E90FF}0.659 & \cellcolor[HTML]{ADD8E6}0.453 & \cellcolor[HTML]{ADD8E6}0.382\\
& \textit{min} & 0.738 & \cellcolor[HTML]{ADD8E6}0.592 & \cellcolor[HTML]{ADD8E6}0.540 & 0.639 & \cellcolor[HTML]{1E90FF}0.387 & 0.544 & \cellcolor[HTML]{1E90FF}0.286 & \cellcolor[HTML]{ADD8E6}0.249 & \cellcolor[HTML]{ADD8E6}0.173\\
& \textit{Original} & 0.799 & 0.679 & 0.656 & 0.709 & 0.467 & 0.643 & 0.521 & 0.364 & 0.219\\
& \textit{avg±var} & 0.824±0.001 & 0.756±0.003 & 0.666±0.003 & 0.713±0.001 & 0.547±0.006 & 0.624±0.001 & 0.514±0.008 & 0.329±0.002 & 0.279±0.002\\
\hline
\multirow{4}{4em}{Llama 3.1 (8b)} & \textit{max} & \cellcolor[HTML]{1E90FF}0.909 & \cellcolor[HTML]{1E90FF}0.898 & \cellcolor[HTML]{1E90FF}0.748 & 0.769 & \cellcolor[HTML]{1E90FF}0.660 & \cellcolor[HTML]{ADD8E6}0.735 & \cellcolor[HTML]{ADD8E6}0.584 & \cellcolor[HTML]{ADD8E6}0.466 & \cellcolor[HTML]{ADD8E6}0.409\\
& \textit{min} & \cellcolor[HTML]{1E90FF}0.376 & \cellcolor[HTML]{1E90FF}0.300 & \cellcolor[HTML]{1E90FF}0.435 & 0.629 & \cellcolor[HTML]{1E90FF}0.166 & \cellcolor[HTML]{ADD8E6}0.456 & \cellcolor[HTML]{ADD8E6}0.320 & \cellcolor[HTML]{ADD8E6}0.252 & \cellcolor[HTML]{ADD8E6}0.175\\
& \textit{Original} & 0.809 & 0.759 & 0.660 & 0.749 & 0.494 & 0.676 & 0.480 & 0.426 & 0.237\\
& \textit{avg±var} & 0.832±0.009 & 0.805±0.011 & 0.668±0.003 & 0.737±0.001 & 0.496±0.013 & 0.652±0.003 & 0.488±0.003 & 0.387±0.002 & 0.282±0.002\\
\hline
\multirow{4}{4em}{Flan-T5-xl (3b)} & \textit{max} & 0.940 & 0.960 & \cellcolor[HTML]{ADD8E6}0.809 & \cellcolor[HTML]{1E90FF}0.776 & \cellcolor[HTML]{ADD8E6}0.743 & \cellcolor[HTML]{1E90FF}0.706 & \cellcolor[HTML]{1E90FF}0.761 & \cellcolor[HTML]{1E90FF}0.685 & \cellcolor[HTML]{1E90FF}0.496\\
& \textit{min} & 0.836 & 0.807 & \cellcolor[HTML]{ADD8E6}0.576 & \cellcolor[HTML]{1E90FF}0.449 & \cellcolor[HTML]{ADD8E6}0.487 & \cellcolor[HTML]{1E90FF}0.166 & \cellcolor[HTML]{1E90FF}0.281 & \cellcolor[HTML]{1E90FF}0.358 & \cellcolor[HTML]{1E90FF}0.155\\
& \textit{Original} & 0.939 & 0.939 & 0.746 & 0.631 & 0.649 & 0.467 & 0.381 & 0.493 & 0.311\\
& \textit{avg±var} & 0.899±0.001 & 0.901±0.002 & 0.695±0.004 & 0.661±0.009 & 0.626±0.003 & 0.539±0.012 & 0.520±0.010 & 0.507±0.008 & 0.328±0.008\\
\hline
\end{tabular}
}
\caption{The maximum (\textit{max}), minimum (\textit{min}), average (\textit{avg}), variance (\textit{var}) of the model performance across different label names in the prompt for each validation set. The performance of the original label set (\textit{Original}) is also included, showing that they fail to reach the maximum performance LLMs could get. The extent of the gap between the maximum and minimum performances is represented using colors: \colorbox[HTML]{1E90FF}{\textit{max}$-$\textit{min}$>0.3$}, \colorbox[HTML]{ADD8E6}{$0.2<$\textit{max}$-$\textit{min}$<0.3$}. The greater the variance, the greater the impact of label lexical choice, and the greater the potential utility of optimizing the label set.}
\label{tab:label name results}
\end{table*}

\paragraph{Prompt Template}

We refer to the prompt template used in the supervised fine-tuning of Flan-T5 and Llama \cite{wang-etal-2022-super,chung2024scaling,dubey2024llama}. We present the results with the following prompt template in this paper: \textit{Given a [text1\_name] and a [text2\_name], detect the stance that the [text2\_name] has towards the [text1\_name]. There are \{N\} options: ``\{label$_0$\}", ``\{label$_1$\}, ... , and \{label$_{N-1}$\}". Now complete the following example. [text1\_name]: \{text1\}. [text2\_name]: \{text2\}}. We also test other templates and find the results are consistent (e.g., prompting with label explanations in Appendix \ref{app:label explain}).

\paragraph{Evaluation}

We adopt macro-$F1$ for model performance evaluation to align with prior studies \cite{schiller2021stance,hardalov-etal-2021-cross,chen-etal-2023-close}. We use $wF2$\footnote{$wF2$ gives different weights for each stance: \textit{deny} $=$ \textit{support} $= 0.40$, \textit{query} $= 0.15$ and \textit{comment} $= 0.05$.} to account for data imbalance in rumoureval \cite{scarton-etal-2020-measuring}.

\paragraph{Implementation Details}

To ensure reproducibility, we use greedy search for decoding\footnote{Potential impact of the decoding strategy can be found in the Appendix \ref{app:decoding}.}. In more than 95\% cases, LLMs exactly follow the instruction and output the stance name within the required stance options. Therefore, we directly use the model generation as the predicted label without post-processing or mapping. We run experiments on the validation set of each dataset. See Appendix \ref{app:label pool} for details on the label sets experimented with. We exclude the topic/entities-based (such as stance towards Obama) stance classification datasets (i.e. scd, semeval and vast) in label elaboration experiments to avoid unnecessary ambiguity or bias during elaboration (e.g., the text to be classified may refer to anything from policy to personal behavior about Obama, and elaborating \textit{agree} to \textit{agree with the opinion of Obama} could lead to biased prediction).

\subsection{Results and Analysis} \label{sec:results analysis}

We first discuss the impact of lexical choice, label order, and label elaboration on zero-shot ICL for classification. Then we provide suggestions for practitioners in zero-shot ICL for classification.

%\begin{figure}[t!]
%\centering

%\begin{subfigure}{0.23\textwidth}
%\includegraphics[width=1\linewidth]{image/llama_order_bestk.png}
%\caption{top-k optimal label sets}\label{subfig:topk-best}
%\end{subfigure}
%\begin{subfigure}{0.23\textwidth}
%\includegraphics[width=1\linewidth]{image/llama_order_worstk.png}
%\caption{top-k poor label sets}\label{subfig:topk-worst}
%\end{subfigure}
%\caption{The max performance gain (positive value) and drop (negative value) on each validation set after re-ordering the label names for the top-k optimal (a) and poor (b) label sets with Llama 3. Results on Flan-T5-xl and Llama 3.1 can be found in the Appendix.}
%\label{fig:label order llama3}
%\end{figure}

\paragraph{Lexical Choice} As shown in Table \ref{tab:label name results}, performance varies across datasets and models solely due to changes in the label names within the prompt. The variations are more pronounced than those reported in previous studies \cite{mu-etal-2024-navigating}. The gap between the highest and lowest performance exceeds 0.1 for all datasets and models, and surpasses 0.2 on more than half of the datasets. We observe that certain stance labels could potentially trigger biased predictions, leading to extremely low performance. For example, Flan-T5 tends to always output \textit{support} when using \textit{support}, \textit{deny}, \textit{neither} for the semeval dataset, and Llama 3 overly predicts \textit{con} when prompted by the label set \textit{pro}, \textit{con}, \textit{neutral} for the emergent dataset.\\

\paragraph{Label Order} On average, we observe limited influence of label order (see the Appendix \ref{app:label order avg} for details). However, we are also interested in extremes of performance change caused by shifting label orders, particularly the maximum performance gain and drop, and whether the extreme gain or drop correlates with the lexicons used for the label names. Therefore, for each dataset, we first select the top-k\footnote{Due to the exponential increasing of label order options and our limited computational resources, we set k=15 for binary classification; k=10 for three-way classification and k=2 for four-way classification} optimal and poor label sets based on their performance in the same order (i.e. the default order). We then examine the maximum increase or decrease of the performance after re-arranging the label orders for optimal and poor label sets, respectively.

%As shown in Figure \ref{subfig:topk-best}, 
We find that altering the order for the optimal label sets has the risk of high performance drops (e.g., even more than 0.2 on the binary classification snopes dataset, Figure \ref{subfig:topk-best} in Appendix). While performance improvements are possible, the gains are relatively limited (lower than 0.1 on all datasets). Conversely, re-arranging the order for the poorly performing label sets offers potential for substantial improvement (Figure \ref{subfig:topk-worst}). This high improvements for certain label sets after re-ordering suggests that the poor performance may partly stem from the initial sub-optimal ordering. 

\paragraph{Label Elaborations} 

Similarly, we select the top-k\footnote{k=15 for binary classification, k=30 for three/four-way classification.} optimal and sub-optimal single-word label sets based on their performance, and examine the performance change after elaborations. 
%As shown in Table \ref{tab:elaboration}, 
We observe that the models are robust to the elaborations for either optimal or poor single-name label sets (full results in Table \ref{tab:apendix label length avg results}), indicating that adding task related details or increasing the label token lengths brings limited impact on average. However, we also observe relatively large performance change on certain datasets. Specifically, for the rumoureval dataset with Llama 3, we notice a performance drop larger than 0.2 when elaborating an optimal label set. Performance on the ibmcs dataset with Llama 3.1 could increase 0.2 when elaborating a poorly-performing label set.

%\begin{table}[h!]
%\centering
%\scalebox{0.6}{
%\begin{tabular}{l|cc|cc|cc}
%\hline
%\multirow{2}{4em}{\textbf{Dataset}} & \multicolumn{2}{c|}{$\mathbf{E_1}$} & \multicolumn{2}{c|}{$\mathbf{E_2}$} & \multicolumn{2}{c}{$\mathbf{E_3}$} \\
%\cline{2-7}
%& \textbf{Opt.} & \textbf{Sub-opt.} & \textbf{Opt.} & \textbf{Sub-opt.} & \textbf{Opt.} & \textbf{Sub-opt.}\\
%\hline
%perspectrum &  0.016 & 0.018 & 0.010 & 0.015 & 0.017 & 0.009\\
%ibmcs & 0.027 & 0.041 & 0.024 & 0.014 & 0.029 & 0.044\\
%snopes & 0.055 & 0.040 & 0.054 & 0.026 & 0.029 & 0.018\\
%emergent & 0.051 & 0.053 & 0.047 & 0.038 & 0.022 & 0.040\\
%rumoureval & 0.095 & 0.036 & 0.084 & 0.020 & 0.058 & 0.102\\
%arc & 0.017 & 0.024 & 0.015 & 0.021 & 0.035 & 0.049\\
%\hline
%\end{tabular}
%}
%\caption{The average absolute performance change after elaborating for \textit{optimal} (\textit{Opt.}) or \textit{poor} (\textit{Sub-opt.}) label sets with Llama 3 ($E_1$, $E_2$, $E_3$ see Figure \ref{fig:overview}). Results on Flan-T5-xl and Llama 3.1 can be found in the Appendix.}  
%\label{tab:elaboration}
%\end{table}

\begin{table*}[ht!]
\centering
\scalebox{0.6}{
\begin{tabular}{l|ccccccccc|cc}
\hline
\multirow{2}{5em}{\textbf{Model}} & \multicolumn{9}{c|}{\textbf{Stance Classificaion}} & \multicolumn{2}{c}{\textbf{Topic Classification}} \\
\cline{2-12}
 & \textbf{perspectrum} & \textbf{ibmcs} & \textbf{snopes} & \textbf{scd} & \textbf{emergent} & \textbf{semeval} & \textbf{rumoureval} & \textbf{arc} & \textbf{vast} & \textbf{TweetTopic} & \textbf{AG News}\\
\hline
Llama 3 (8b) & -0.4921* & -0.3787* & -0.4359* & -0.4217* & -0.5781 & -0.2618* & -0.3764* & -0.1662 & -0.3639* & -0.4994* & -0.2447*\\
\hline
Llama 3.1 (8b) & -0.4103* & -0.3642* & -0.1874 & -0.0686 & -0.4310* & -0.2232* & -0.3944* & -0.1708* & -0.1208 & -0.2196* & 0.0200\\
\hline
Flan-T5-xl (3b) & -0.4476* & -0.3638* & -0.6014* & 0.2353 & -0.1089 & -0.2881* & -0.1714* & -0.5742 & 0.1003 & 0.1587 & 0.0398\\
\hline
\end{tabular}
}
\caption{Spearman correlation co-efficiency between model performance on validation set and kurtosis of neurons in the last layer. Mark with * when p value is lower than 0.05.}
\label{tab:Spearman correlation}
\end{table*}

\subsection{Suggestions to Practitioners}\label{subsec:suggestions}
Based on our results and analysis, we provide the following suggestions to practitioners in zero-shot ICL for classification:

\begin{enumerate}[leftmargin=*, noitemsep]
    \item Lexical designing for the label names should be considered as an important step in prompt engineering. 
    \item Single-word class label without elaboration on task information is able to achieve high performance in most cases, i.e. adding extra information does not yield better results.
    \item If the practitioner has selected a set of optimal lexicons for label options based on a specific order, exploring alternative label orders can be redundant due to the limited performance gain brought from high computational costs. However, if a label set is chosen randomly, experimenting with different label orders may yield meaningful improvements (see Figure \ref{fig:app label order}).
\end{enumerate}

\section{Neuron Analysis for Label Selection} \label{sec:neuron}

Although our findings indicate the importance of label word selection for text classification in zero-shot ICL, current studies lack consideration of this factor. Therefore, we conduct empirical analysis to gain insights into the underlying mechanism of lexical choice for single-word label names. 

We preliminarily explored related approaches discussed in Section \ref{sec:related work}, including prompt perplexity~\cite{gonen-etal-2023-demystifying} and model internal representation of label words~\cite{wang-etal-2023-label}, but they did not yield any significant correlation with zero-shot ICL model performance (see the Appendix \ref{app:ppl} and \ref{app:labelkey} for details). Meanwhile, various studies have indicated the correlation between neuron activation pattern in FFN and model performance \cite{kuzmin2023pruning,tang-etal-2024-language,stolfo2024confidence,wu-etal-2024-language}. Inspired by the finding that the presence of outliers in neural networks is predictive of quantization and pruning performance for the layers of LLMs \cite{kuzmin2023pruning}, %Inspired by studies on pre-trained language models or LLMs that indicate the correlation between neuron activation pattern in FFN and model performance \cite{kuzmin2023pruning,tang-etal-2024-language,stolfo2024confidence,wu-etal-2024-language}, 
we establish a new hypothesis: the model performance influenced by label names is correlated with the number of outliers in the neurons within FFN in the decoder of the LLMs. We empirically validate our hypothesis on the nine stance classification datasets, as well as two topic classification datasets (AG News \cite{zhang2015character} and TweetTopic \cite{antypas-etal-2022-twitter}) to show the generalizability to other NLP tasks.

\paragraph{Methodology} For each FFN module in layer $i$ in the decoder, it can be denoted as follows: 
\begin{equation}
h^i = (\text{act\_fn}(\tilde{h}^i W^i_1) \otimes \tilde{h}^i W^i_3) \cdot W^i_2.
\end{equation}

\noindent{where $\tilde{h}^i$ is the output hidden states from multi-head self-attention module. The activation function ($\text{act\_fn}$) for Flan-T5 and Llama 3/3.1 is Gaussian Error Linear Unit (GELU) \cite{hendrycks2016gaussian} and Sigmoid Linear Unit (SiLU) \cite{hendrycks2016gaussian,elfwing2018sigmoid}, respectively.} 

A \textit{neuron} is defined as the linear transformation of each column in $W^i_1$ followed by the activation function. Here, we study the last layer $I$'s output of $\text{act\_fn}(\tilde{h}^I W^I_1)$ (denoted as $N_I$) for the predicted first token of the label name in the model generation. Following Kuzmin et al.,~\shortcite{kuzmin2023pruning}, we measure the number of outliers over the neuron output distribution ($N_I$) through kurtosis, given by:
\begin{equation}
\text{Kurtosis}[N_I] =\frac{\mathbb{E}[(N_I - \mu)^4]}{(\mathbb{E}[(N_I - \mu)^2])^2} 
\end{equation}
where $\mu$ is the mean of $N_I$. 
For each dataset, we average the kurtosis scores over the validation set for each candidate label set. We then calculate the Spearman correlation between model performance and the averaged kurtosis score.

\paragraph{Results} 

Table \ref{tab:Spearman correlation} shows that, for most datasets, there is statistically significant negative correlation between model performances and kurtosis scores across models and activation functions, indicating that fewer outliers in the neurons of the final layers are associated with enhanced zero-shot ICL performance. 
%We visualize an example of such correlation in Figure \ref{fig:correlation example}. 
This observation implies that the kurtosis score of neuron activation distribution in FFN of the last decoder layer of LLMs could potentially serve as an effective signal for selecting optimal label names in zero-shot classification.

%\begin{figure}[h!]
 %\centering
 %\includegraphics[width=0.4\textwidth]{image/correlation.png}
 %\caption{Macro-F1 vs. neuron activation kurtosis for the emergent dataset with different label options in the prompt for Llama 3.} 
 %\label{fig:correlation example}
 % \cass{maybe color "endorse" and "for" in red, and "deny" and "against" in blue, etc}}
%\end{figure}

\begin{table*}[ht!]
\centering
\scalebox{0.6}{
\begin{tabular}{l|l|ccccccccc|cc}
\hline
\multirow{2}{5em}{\textbf{Model}} & \multirow{2}{5em}{\textbf{Method}} & \multicolumn{9}{c|}{\textbf{Stance Classification}} & \multicolumn{2}{c}{\textbf{Topic Classification}} \\
\cline{3-13}
 &  & \textbf{perspectrum} & \textbf{ibmcs} & \textbf{snopes} & \textbf{scd} & \textbf{emergent} & \textbf{semeval} & \textbf{rumoureval} & \textbf{arc} & \textbf{vast} & \textbf{TweetTopic} & \textbf{AG News }\\
\hline
\multirow{4}{5em}{Llama 3 (8b)} & \textit{\loads{}} & \underline{0.8431} & \underline{0.7684} & \underline{0.5516} & \underline{0.6698} & \underline{0.5870} & \underline{0.6445} & \underline{0.4097} & \underline{0.3662} & \underline{0.3190} &\underline{0.7752} & \underline{0.7660} \\
& \textit{Original Label} & 0.8187 & 0.5485 & 0.4387 & 0.6504 & 0.3416 & 0.5565 & 0.3487 & 0.3130 & 0.2395 & 0.7742 & 0.7594\\
& \textit{Original + Verbalizer} & 0.8185 & 0.5485 & 0.4306 & 0.6578 & 0.3400 & 0.5576 & 0.3476 & 0.3131 & 0.2395 & 0.7742 & 0.7594\\
& \textit{Self-generated} & 0.6912 & 0.5802 & 0.3314 & 0.6607 & 0.3692 & 0.6032 & 0.2745 & 0.2837 & 0.3070 & 0.5851 & 0.6235\\
\hline
\multirow{4}{5em}{Llama 3.1 (8b)} & \textit{\loads{}} & \underline{0.8789} & \underline{0.8856} & \underline{0.6212} & \underline{0.7274} & \underline{0.6403} & \underline{0.6581} & \underline{0.3642} & 0.3637 & 0.2458 & \underline{0.7988} & \underline{0.7306} \\
& \textit{Original Label} & 0.8064 & 0.6783 & 0.4426 & 0.6983 & 0.4337 & 0.6492 & 0.3528 & 0.3784 & 0.2126 & 0.7909 & 0.7273\\
& \textit{Original + Verbalizer} & 0.8064 & 0.6783 & 0.4426 & 0.6983 & 0.4262 & 0.6448 & 0.3534 & \underline{0.3838} & 0.2128 & 0.7909 & 0.7273\\
& \textit{Self-generated} & 0.6244 & 0.4918 & 0.5146 & 0.6798 & 0.2984 & 0.6421 & 0.3409 & 0.3373 & \underline{0.2578} & 0.5956 & 0.6217\\
\hline
\multirow{4}{5em}{Gemma 2 (9b)} & \textit{\loads{}} & \underline{0.9110} & \underline{0.9247} & 0.6233 & 0.7595 & \underline{0.5989} & 0.6784 & \underline{0.4896} & \underline{0.4858} & \underline{0.3439} & 0.8032 & \underline{0.8451} \\
& \textit{Original Label} & 0.8966 & 0.8728 & 0.6454 & \underline{0.7707} & 0.5704 & \underline{0.6874} & 0.4778 & \underline{0.4858} & 0.3383 & 0.8241 & 0.8316\\
& \textit{Original + Verbalizer} & 0.8966 & 0.8728 & 0.6424 & \underline{0.7707} & 0.5523 & 0.6873 & 0.4689 & 0.4748 & 0.3383 & \underline{0.8246} & 0.8322\\
& \textit{Self-generated} & 0.9017 & 0.9113 & \underline{0.6656} & 0.7480 & 0.5833 & 0.6621 & 0.3549 & 0.4657 & 0.3178 & 0.6303 & 0.8215\\
\hline
\multirow{4}{5em}{Flan-T5-xl (3b)} & \textit{\loads{}} & \underline{0.9334} & \underline{0.9380} & 0.6881 & 0.5997 & \underline{0.5813} & 0.4951 & \underline{0.4759} & \underline{0.4688} & \underline{0.3857} & \underline{0.8530} & 0.8556\\
& \textit{Original Label} & 0.9305 & 0.8971 & \underline{0.7267} & 0.6341 & 0.5580 & 0.5697 & 0.3837 & 0.4628 & 0.3473 & 0.8071 & \underline{0.9209}\\
& \textit{Original + Verbalizer} & 0.9305 & 0.8953 & 0.7026 & 0.6507 & 0.5586 & 0.5712 & 0.3852 & 0.4613 & 0.3482 & 0.8091 & \underline{0.9209}\\
& \textit{Self-generated} & 0.7914 & 0.7384 & 0.3677 & \underline{0.6974} & 0.3883 & \underline{0.5954} & 0.2986 & 0.3343 & 0.3042 & 0.8501 & 0.7368\\
\hline
\multirow{4}{5em}{Flan-T5-xxl (13b)} &  \textit{\loads{}} & \underline{0.9428} & \underline{0.9644} & 0.6905 & 0.6158 & \underline{0.5938} & 0.5257 & \underline{0.3852} & \underline{0.6151} & \underline{0.4218} & \underline{0.7727} & 0.7742\\
& \textit{Original Label} & 0.9407 & 0.9630 & \underline{0.7814} & \underline{0.7598} & 0.5614 & 0.5697 & 0.2016 & 0.6065 & 0.3278 & 0.6643 & \underline{0.9177}\\
& \textit{Original + Verbalizer} & 0.9407 & 0.9621 & 0.7716 & 0.7598 & 0.5631 & 0.5705 & 0.2011 & 0.6062 & 0.3278 & 0.6643 & \underline{0.9177}\\
& \textit{Self-generated} & 0.8622 & 0.8384 & 0.4226 & 0.7315 & 0.4309 & \underline{0.6182} & 0.3110 & 0.6115 & 0.2881 & 0.7242 & \underline{0.9177}\\
\hline
\end{tabular}
}
\caption{Comparison of zero-shot ICL performance on test sets between prompting with \loads{}-selected label names versus the other three baseline approaches. We underline the highest model performance (statistically significant with paired chi-squared test).}
\label{tab:Performance comparison}
\end{table*}

\section{\loads{}: Label set Optimization via Activation Distribution kurtosiS} \label{sec:load}

Motivated by the above observation that the fluctuated zero-shot ICL performance caused by different label names could be attributed to the number of outliers in neurons in the last layer of LLMs, we propose \loads{} to obtain an optimal label set for a given classification task in a post-hoc setting. 
% Specifically, we design the following three-step pipeline based on \loads{} for automatic label selection in zero-shot ICL:

\subsection{Method}

We design a three-step pipeline based on \loads{} for automatic label selection in zero-shot ICL:
\begin{enumerate}[leftmargin=*, noitemsep]
%\small
    \item Create a list of candidate label sets for class options in the prompt (see Section \ref{sec:label setup}). The label names in each set should follow the same order.
    \item Rank the list of label options based on the kurtosis score of the neuron activation in FFN of the last decoder layer (averaged across the validation set).
    \item Choose the label set with the lowest averaged kurtosis score.
\end{enumerate}
The above \loads{}-selected label set can then be used in the standard zero-shot ICL on test sets.

\subsection{Evaluation}

\paragraph{Setups}

We randomly sample 100 data points from the validation set for label selection and test the selected label sets on the official test sets. 
\begin{itemize}[leftmargin=*, noitemsep]
    \item \textbf{Baselines:} We compare the model performance of using label words selected by \loads{} to the following three approaches: (1) \textit{Original label words}: we use the original label words from the dataset (i.e., the labels in Table \ref{tab:datasets}), as it is the conventional and widely adopted practice; (2) \textit{Original label words with a verbalizer}: after prompting the LLMs with the original label words, we employ our pool of candidate label words (See Section \ref{sec:Lexical Choice}) as a verbalizer and incorporate their probabilities into the final probability of the same class; (3) \textit{Self-generated label words}: we prompt the LLMs without providing any class options and select the candidate label words with the average highest probability at the first generated label token.

    \item \textbf{LLMs:} In addition to \textit{Llama3} (8b), \textit{Llama 3.1} (8b) and \textit{Flan-T5-xl} (3b), we also examine whether \loads{} could generalize to other model families and model sizes by including instruction-tuned \textit{Gemma-2-it (9b)} and \textit{Flan-T5-xxl} (13b).

    % \item \textbf{Datasets:} To test whether \loads{} could generalise to NLP tasks beyond stance classification, we explore \textit{topic classification}, a task involving various topic categories across different datasets. Given that topic classification is commonly included in supervised fine-tuning (SFT) datasets (e.g., T0-SF \cite{sanh2022multitask}) widely used by LLMs such as Flan-T5 models, we consider AG News \cite{zhang2015character} (in T0-SF)  and TweetTopic \cite{antypas-etal-2022-twitter} (not in any public SFT datasets)  to explore the impact of potential data leakage on \loads{}.
    
\end{itemize}

\paragraph{Results}

Table \ref{tab:Performance comparison} presents the zero-shot ICL model performances on stance classification and topic classification datasets with different label word selection strategies. The results demonstrate that employing \loads{} to select label sets for zero-shot ICL prompts yields superior performance compared to other baseline approaches on most of the datasets. The improvement is consistent across NLP tasks and datasets, model architectures and sizes, as well as prompt templates\footnote{The analysis of prompt sensitivity can be found in the Appendix \ref{app:loads_sen}.}. 

Also, we observe limited benefits from adopting the verbalizer in post-processing, since LLMs tend to give the predicted label word tokens high probability in most of cases. Prompting with the self-generated label words rarely results in the best performance, while with the risk of leading to extremely low performance on certain datasets (e.g., perspectrum and snopes datasets with Llama 3).

Furthermore, potential data leakage could have significant impact on \loads{}, indicated by the high performance achieved by the original label words on the AG News dataset with Flan-T5 models which was instruction-tuned with this dataset. It also aligns with the results in Table \ref{tab:Spearman correlation} where no statistically negative correlation was observed on the AG news dataset with Flan-T5-xl.

\subsection{Analysis}

\paragraph{Cross-lingual Transferability}

Previous research \cite{zhang-etal-2023-dont} has demonstrated that for non-English datasets, prompting with English task instructions (including label words) while keeping the input in the original language often yields superior performance than non-English instructions. Therefore, we investigate whether the optimal English label sets selected by \loads{} on English dataset can also enhance performance for non-English datasets in this scenario.

We manually translate the English rumoureval Twitter test set into French and Portuguese. We select the optimal label set based on \loads{} with 100 randomly sampled data from the English rumoureval validation set. We then prompt the LLMs with English task instructions and French/Portuguese inputs. Table \ref{tab:cross-lingual} presents the results, indicating that the optimal English label set identified by \loads{} also effectively improves performance on non-English datasets when the instruction is provided in English.

\begin{table}[h!]
\centering
\scalebox{0.7}{
\begin{tabular}{l|l|cc}
\hline
\textbf{Model} & \textbf{Method} & \textbf{French} & \textbf{Portuguese}\\
\hline
\multirow{2}{5em}{Llama 3 (8b)} & \textit{\loads{}} & \underline{0.5020} & \underline{0.4528}\\
& \textit{Original Label} & 0.4544 & 0.4284\\
\hline
\multirow{2}{5em}{Llama 3.1 (8b)} & \textit{\loads{}} & \underline{0.4728} & \underline{0.4278}\\
& \textit{Original Label} & 0.3731 & 0.3679\\
\hline
\multirow{2}{5em}{Flan-T5-xl (3b)} & \textit{\loads{}} & \underline{0.5137} & \underline{0.3912}\\
& \textit{Original Label} & 0.4189 & 0.3317\\
\hline
\end{tabular}
}
\caption{Performance when LLMs are prompted with English instructions (including label options) and French/Portuguese inputs. Label options are selected by \loads{} with English validation data.}
\label{tab:cross-lingual}
\end{table}

\paragraph{Data Efficiency}

To show the merit of data efficiency of \loads{}, we randomly sample 50, 100, 300, 500 or 1000 data points from the validation set and compare the rankings of the label sets based on \loads{}. Due to the resource restriction, we conduct experiments on snopes (binary classification) and emergent (three-way classification) datasets with Flan-T5-xl and Llama 3.

The results show that the rankings of the top 5 label sets remain consistent across different sample sizes. It suggests that \loads{} can achieve comparable performance even with a smaller number of unlabeled data samples than 100, further highlighting the data efficiency of our proposed method.

% \paragraph{Alternative Methods} We explore other approaches to improve \loads{}, including prompt perplexity~\cite{gonen-etal-2023-demystifying} (see Appendix \ref{app:ppl}) and model internal representation of label names~\cite{wang-etal-2023-label} (see Appendix \ref{app:labelkey}), but they did not yield any significant correlation with zero-shot ICL model performance.

\paragraph{Label Transferability} 

%A natural question arising from the application of \loads{} in practice might be: \textit{For NLP tasks such as stance classification where different label lexicons are used to represent the same classes across datasets, can \loads{}-selected label sets be generalised across datasets or even models?} 
%If so, for any new datasets/models, we could potentially adopt the optimal label words identified by \loads{} on other related datasets/models without collecting samples for the target new datasets or conducting additional experiments on new LLMs. 
We explore whether \loads{}-selected label sets can be generalised across datasets or even models for NLP tasks such as stance classification where different label lexicons are used to represent the same classes across datasets. Specifically, for each dataset $D_i$ on each LLM $M_j$, we select the optimal label set $L_{D_iM_j}$ through \loads{}.
% , and calculate the overlap between the label sets. Specifically, 
To analyse whether the \loads{}-selected label set on one dataset could be adapted to another related dataset with the same LLM $M$, we calculate the overlap of optimal labels ($L_{D_i\_M}$) between each dataset. Similarly, we examine the overlap of optimal labels ($L_{D\_M_i}$) between each model to explore whether the \loads{}-selected optimal labels for dataset $D$ could be adapted across LLMs. We only focus on the positive and negative stances to enable comparison across binary, three-way and four-way stance classification datasets.

Our results indicate that \loads{}-selected label sets is transferable across datasets on the same LLM, highlighting the potential of leveraging \loads{} to identify optimal label words with established related datasets, avoiding the need to collect samples for the target new dataset. For example, the positive-negative stance label pairs identified for Llama 3 is \textit{endorse} and \textit{deny} across all the stance classification datasets. However, we find that the label words selected for a specific dataset on one LLM often differ from those identified for another LLM, suggesting \loads{}' dependency on the underlying model architectures and parameters.

In summary, the \loads{}-selected label sets tend to be model-dependent rather than dataset-dependent. This observation aligns with the mechanism of \loads{}, as the neurons and their distributions are inherently tied to the specific model. We hypothesize that this may suggest a correlation between the \loads{}-selected label words and the LLMs' internal representation or understanding of the target NLP task or concept (e.g., what is stance), highlighting potential directions for future studies.

\section{Conclusion}

We study the impact of label options in the prompt for classification in zero-shot ICL, including lexical choice, label order, and label elaborations. We observe a significant effect of the lexicons used to represent label words in the prompt, also linking to the models' sensitivity to the label order. 
% This work presents the first comprehensive empirical study investigating the influence of label set design on zero-shot in-context learning performance for stance classification tasks. Our findings reveal that the lexical choice of label words plays a critical role in model performance, while label order and excessive elaboration have less impact. 
Through neuron activation analysis, we find that optimal label sets produce fewer outlier neurons in LLMs' feed-forward networks. We then propose \loads{}, a novel method for selecting optimal label sets using activation distribution kurtosis. Prompting with \loads{}-selected label sets consistently outperforms the use of original dataset labels across different models. Our approach is post-hoc, data-efficient and requires no gradient propagation or model fine-tuning. It also demonstrates cross-lingual transferability when using English instructions for non-English datasets. %Our findings have important implications for leveraging LLMs in zero-shot classification scenarios. 
By showing that carefully selecting label sets based on neuron activation patterns can significantly enhance model performance without requiring additional training or labeled data, this paper has important implications for leveraging LLMs in zero-shot classification. %Our work also opens up new avenues for optimizing prompts and instructions for LLMs in classification tasks.

\section*{Limitations}

Our experiments focused primarily on stance classification tasks. We chose this task because the label ambiguity is an identified challenge (i.e., label names could be replaced by a sufficient number of synonyms without altering their meanings and scopes in the original study), and it has sufficient datasets for empirical study. Although we have tested the generalisabity of our findings on topic classification, with more datasets released and new tasks proposed in future studies,  studies could be conducted to explore whether our findings generalize to a broader range of classification tasks and domains. Also, although we examined multiple models from the Flan-T5 and Llama families, our study did not include other popular language models (such as Phi\footnote{\url{https://huggingface.co/microsoft/phi-2}} or Mistral \footnote{\url{https://huggingface.co/mistralai/Mistral-7B-Instruct-v0.2}}) due to computational resource limitation. Expanding the range of models would provide a more comprehensive understanding of label option's impact across different architectures.

Another limitation of our study is English-language bias. Although we have explored the cross-lingual transferability to French and Portuguese on one dataset, more extensive multilingual testing is needed to ensure the approach's effectiveness across diverse languages and cultures. %While we observed a correlation between neuron activation patterns and model performance, a deeper theoretical analysis is needed to fully explain why certain label sets perform better than others.

Our method, while more efficient than gradient-based approaches, still requires running inference on a subset of data to compute activation statistics. This may be challenging for resource-constrained environments or very large models. The efficiency of our method may also be challenged when the classification task contains a very large number of class categories. Furthermore, we ensure the inclusion of samples for each class. The effectiveness of our method might vary when the validation set is highly imbalanced or even lack of data for the minority class. The effectiveness of different distribution metrics is out of scope but we acknowledge that it may have significant improvement for our method.  

Lastly, while we focused on technical performance, future work should consider potential biases introduced by label set choices and their implications for fairness and inclusivity in classification tasks. Addressing these limitations in future research will help to further validate and refine our approach to optimal label set selection for zero-shot ICL.

\section*{Acknowledgments}

This work is funded by EMIF managed by the Calouste Gulbenkian Foundation\footnote{The sole responsibility for any content supported by the European Media and Information Fund lies with the author(s) and it may not necessarily reflect the positions of the EMIF and the Fund Partners, the Calouste Gulbenkian Foundation and the European University Institute.} under the "Supporting Research into Media, Disinformation and Information Literacy Across Europe" call (ExU -- project number: 291191).\footnote{\url{exuproject.sites.sheffield.ac.uk}} and the UK’s innovation agency (InnovateUK) grant number 10039039 (approved under the Horizon Europe Programme as VIGILANT\footnote{\url{https://www.vigilantproject.eu}}, EU grant agreement number 101073921).

\bibliography{custom,latex/anthology}

\appendix

\section{Datasets}
\label{app:dataset}

We summarise the datasets we used in our study in Table \ref{tab:app_datasets}. For the stance classification datasets without official validation sets, we use the train/validation splits provided by Schiller et al., \shortcite{schiller2021stance}. We utilize the official test set for AG news\footnote{\url{https://huggingface.co/datasets/sh0416/ag_news}}, and the official “train/test random split in the COLING 2022 paper" for TweetTopic dataset\footnote{\url{https://huggingface.co/datasets/cardiffnlp/tweet_topic_single}}. TweetTopic dataset has six class categories, potentially resulting in more than 4,000 different label sets if we consider only five synonymy words for each category (i.e., more than 12,000 experiments on three LLMs). Due to our limited computational resource, we experiment with three topics: \textit{pop culture},
\textit{daily life}, and \textit{science \& technology}.

\begin{table}[h!]
\centering
\scalebox{0.7}{
\begin{tabular}{ll|c}
\hline
Dataset Name & Source & \# of Label Sets\\
\hline
scd & Debates & 31\\
perspectrum  & Debates & 31\\
snopes & News & 31\\
ibmcs & Debates + Wikipedia & 31\\
vast & Debates + Artificial & 62\\
emergent & News & 62\\
semeval & Social Media & 93\\
rumoureval & Social Media & 248\\
arc & Debates & 62\\
\hline
AG News & News & 50\\
TweetTopic & Social Media & 64 \\
\hline
\end{tabular}
}
\caption{Datasets and the number of label sets we experiment with for each dataset.}
\label{tab:app_datasets}
\end{table}

\section{Data Leakage}

As far as we know, Llama 3 and Llama 3.1 are not supervised fine-tuned with any public stance classification datasets. Flan-T5 is fine-tuned on Super-NaturalInstructions dataset \cite{wang-etal-2022-super}, containing two English stance classification tasks \cite{kobbe-etal-2020-unsupervised} (i.e., task 209 and 513). The tasks are formed as a binary (“in favor” and “against”) and a three-way (“in favor”, “against” and “neutral”) classification, respectively. There is no overlapping between these two datasets and our nine experimented datasets.

\section{Label Pool Creation} \label{app:label pool}
 
\paragraph{Stance Classification} Following the pipeline we described in Section 3 of the main paper, we collect the seed label sets from the nine stance classification datasets (see Table 1). For the semeval dataset, the label set in the original paper ("favor, against, neither" in Table 1 in main paper) is slightly different from the set used in their published dataset ("favor, against, none"), so we consider both of them. 

For positive and negative stance label names, we aim to acquire word-pairs with semantically opposite meaning. We first extract antonym for each positive and negative seed stance label from WordNet. Since we obtain limited antonyms in this way, Claude is then used to generate synonym for each seed positive-negative stance label pairs. An example of the prompt we used is: \textit{Provide 5 different pairs of synonyms for "support" and "deny". They are supposed to be labels for stance classification.} We use WordNet to obtain synonyms for the rest of stance labels if there are any. For the label names that represent "neutral" stance in the original study, such as "observing" and "comment", we take "neutral" as their synonyms. Finally, we manually select the appropriate label names generated by Claude. The number of label sets we experiment with for each dataset is listed in Table \ref{tab:app_datasets}.

\paragraph{Topic Classification} Similarly, we follow the pipeline to collect and generate synonyms for each topic category. For TweetTopic dataset, since \textit{pop culture} is a mixture of multiple sub-topics as discussed by Antypas et al.,\shortcite{antypas-etal-2022-twitter}, we also consider the synonyms of the sub-topics. We use every possible combination of synonyms among topic categories for TweetTopic dataset. For AG news, there are total 160 combinations. We randomly sample 50 of them due to limited computational resources. The number of label sets we experiment with for two datasets is listed in Table \ref{tab:app_datasets}.

\section{Decoding Strategies} \label{app:decoding}

We adjust the temperature (0.2, 0.4, 0.6, 0.8, 1.0, 1.2, 1.4) used for sampling-based decoding and compare their performances with the greedy search based performance for emergent and snopes dataset on Flan-T5-xl and Llama 3.

A temperature value larger than 1.0 -- flattening the probability distribution -- tends to harm the performance especially for Flan-T5-xl, which generates outputs irrelevant to stance. When temperature is lower than 1.0, introducing randomness in decoding through sampling may benefit the performance, but not significantly (in most of cases improvement is lower than 0.07). We summarise the maximum performance increase or decrease comparing with greedy search in Table \ref{tab:app_decoding}. 

\begin{table}[h!]
\centering
\scalebox{0.7}{
\begin{tabular}{cc|cccc}
\hline
\multirow{2}{6em}{Model Name} & \multirow{2}{5em}{Temperature} & \multicolumn{2}{c}{snopes} & \multicolumn{2}{c}{emergent}\\
 & & + & - & + & -\\
\hline
\multirow{7}{5em}{Flan-T5} & 0.2 & 0.021 & 0.049 & 0.037 & 0.049\\
& 0.4 & 0.055 & 0.066 & 0.066 & 0.068\\
& 0.6 & 0.033 & 0.099 & 0.034 & 0.064\\
& 0.8 & 0.033 & 0.145 & 0.066 & 0.107\\
& 1.0 & 0.022 & 0.189 & 0.036 & 0.149\\
& 1.2 & 0.037 & 0.264 & 0.046 & 0.196\\
& 1.4 & 0.015 & 0.428 & 0.042 & 0.361\\
\hline
\multirow{7}{5em}{Llama 3} & 0.2 & 0.050 & 0.063 & 0.068 & 0.035\\
& 0.4 & 0.044 & 0.047 & 0.073 & 0.073\\
& 0.6 & 0.040 & 0.062 & 0.075 & 0.059\\
& 0.8 & 0.027 & 0.062 & 0.066 & 0.107\\
& 1.0 & 0.054 & 0.086 & 0.100 & 0.103\\
& 1.2 & 0.038 & 0.072 & 0.069 & 0.171\\
& 1.4 & 0.030 & 0.104 & 0.128 & 0.142\\
\hline
\end{tabular}
}
\caption{The maximum performance increase (+) and decrease (-) if adopting sampling-based decoding rather than greedy search.}
\label{tab:app_decoding}
\end{table}

\section{Prompting with Label Explanation} \label{app:label explain}

We investigate whether the performance variance caused by different lexical choices of the label names could be mitigated or lowered by including the explanation of the label names in the prompt. We experiment with emergent and snopes datasets on Flan-T5-xl and Llama 3. We add the following class explanations in the prompt template after the class options for snopes and emergent datasets respectively: (1) snopes: \textit{If the text supports that claim, answer with "\{positive stance\}"; if the text opposes the claim, answer with "\{negative stance\}"}; (2) emergent: \textit{If the headline supports the claim, answer with "\{positive stance\}"; if the headline opposes the claim, answer with "\{negative stance\}"; if the claim is discussed in the headline but without assessment of its veracity, "\{neutral stance\}"}.

We observe that including these label name explanations in the prompt may help with the label sets that achieve the lowest zero-shot performance. As for the snopes dataset, its worst performance would increase from 0.5400 to 0.555 (Llama 3, labels: \textit{supportive} and \textit{opposed}) or from 0.5766 to 0.6568 (Flan-T5-xl, labels: \textit{for}, \textit{against}). As for emergent, its lowest performance would increase significantly from 0.3877 to 0.5600 with Llama 3 (labels: \textit{pro}, \textit{con} and \textit{neutral}). However, when using Flan-T5-xl, the inclusion of the class explanation even decrease the worst performance from 0.4870 to 0.3775 (labels: \textit{support}, \textit{deny} and \textit{neutral}). 

More importantly, the benefits from label explanations in the prompt would not close the gap between the optimal and sub-optimal label sets, comparing the above improved performance with the maximum performances in Table 2 in the main paper.

\section{Label Order Results} \label{app:label order avg}

We present the averaged absolute performance difference after re-ordering the label names in the prompt in Table \ref{tab:appendix label order results}. The influence is limited on average. 

\begin{table}[h!]
\centering
\scalebox{0.7}{
\begin{tabular}{lccc}
\hline
Dataset & Flan-T5 & Llama 3 & Llama 3.1 \\
\hline
perspectrum & 0.0148 & 0.0372 & 0.0771\\
ibmcs & 0.0195 & 0.0696 & 0.0961\\
snopes & 0.0296 & 0.0772 & 0.1259\\
scd & 0.0309 & 0.0288 & 0.0484\\
emergent & 0.0211 & 0.0689 & 0.1578\\
semeval & 0.0230 & 0.0304 & 0.0527\\
vast & 0.0311 & 0.0465 & 0.0495\\
rumoureval & 0.0355 & 0.0720 & 0.0439\\
arc & 0.0152 & 0.0606 & 0.0612\\
\hline
\end{tabular}
}
\caption{The average absolute performance change after re-ordering the label options in the prompt.}
\label{tab:appendix label order results}
\end{table}

\begin{figure*}
\centering
\begin{subfigure}{0.35\textwidth}
\includegraphics[width=1\linewidth]{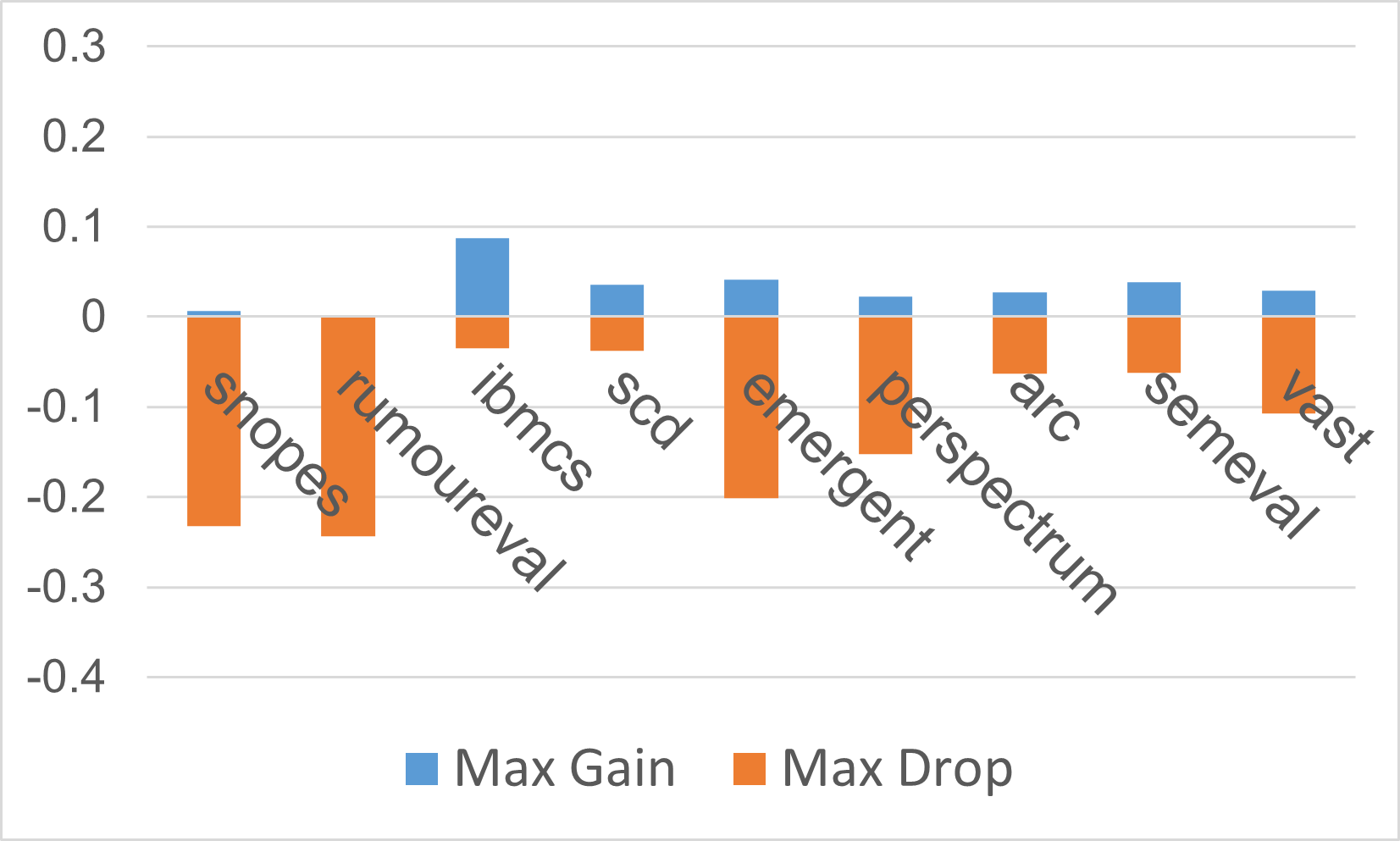}
\caption{Llama3: top-k optimal label sets}\label{subfig:topk-best}
\end{subfigure}
\begin{subfigure}{0.35\textwidth}
\includegraphics[width=1\linewidth]{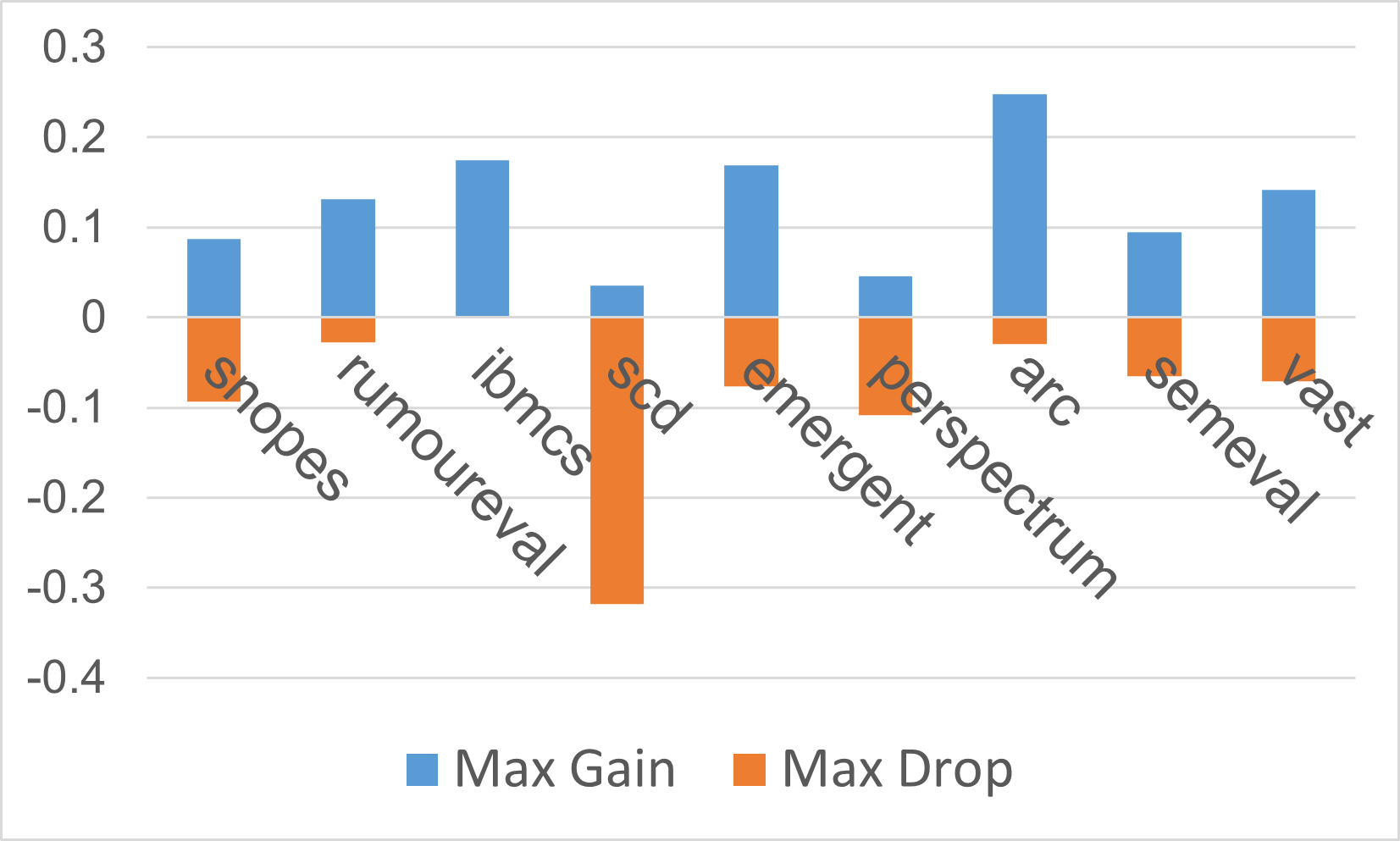}
\caption{Llama3: top-k sub-optimal label sets}\label{subfig:topk-worst}
\end{subfigure}
\begin{subfigure}{0.35\textwidth}
\includegraphics[width=1\linewidth]{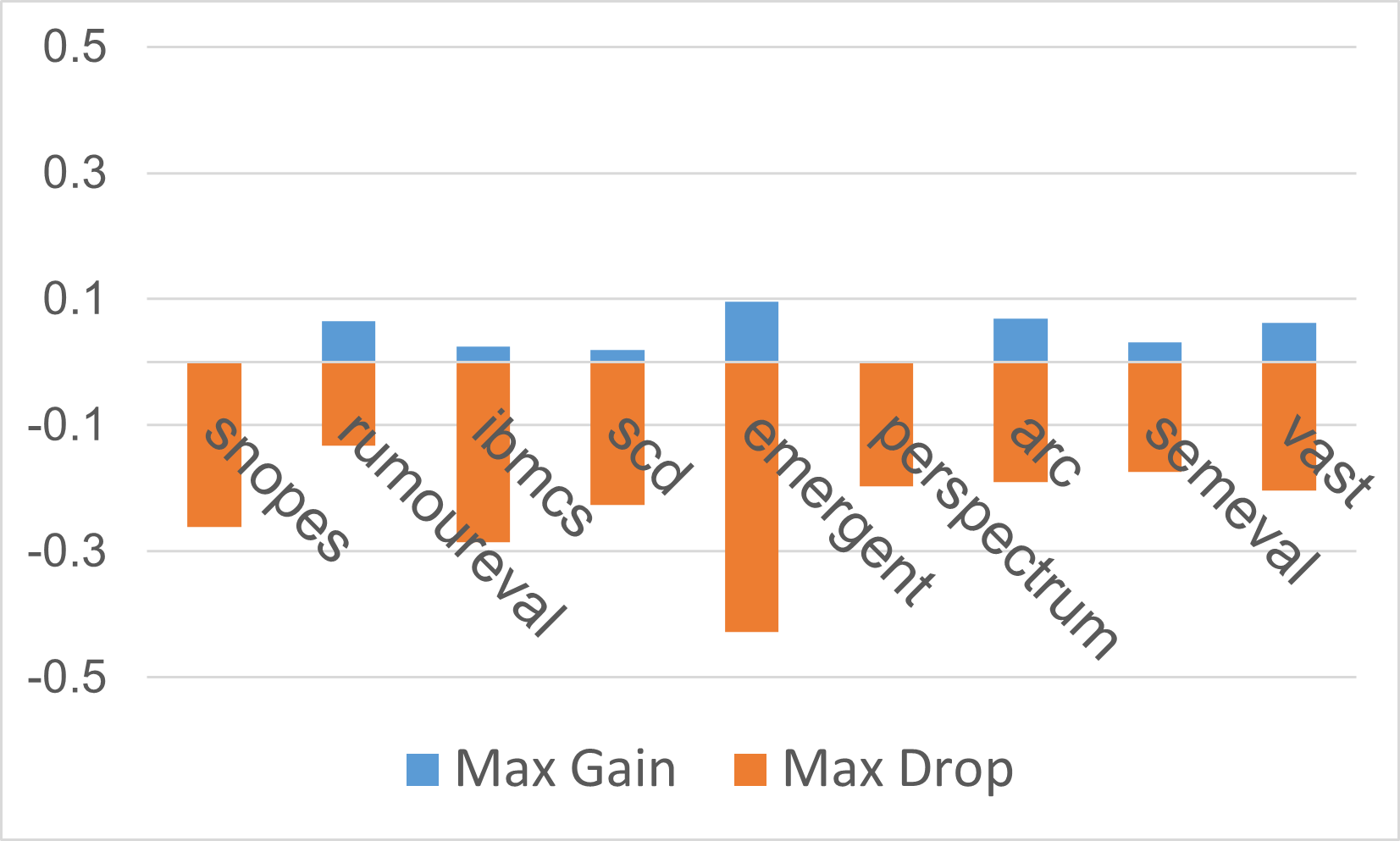}
\caption{Llama 3.1: top-k optimal label sets}\label{subfig:llama31_topk-best}
\end{subfigure}
\begin{subfigure}{0.35\textwidth}
\includegraphics[width=1\linewidth]{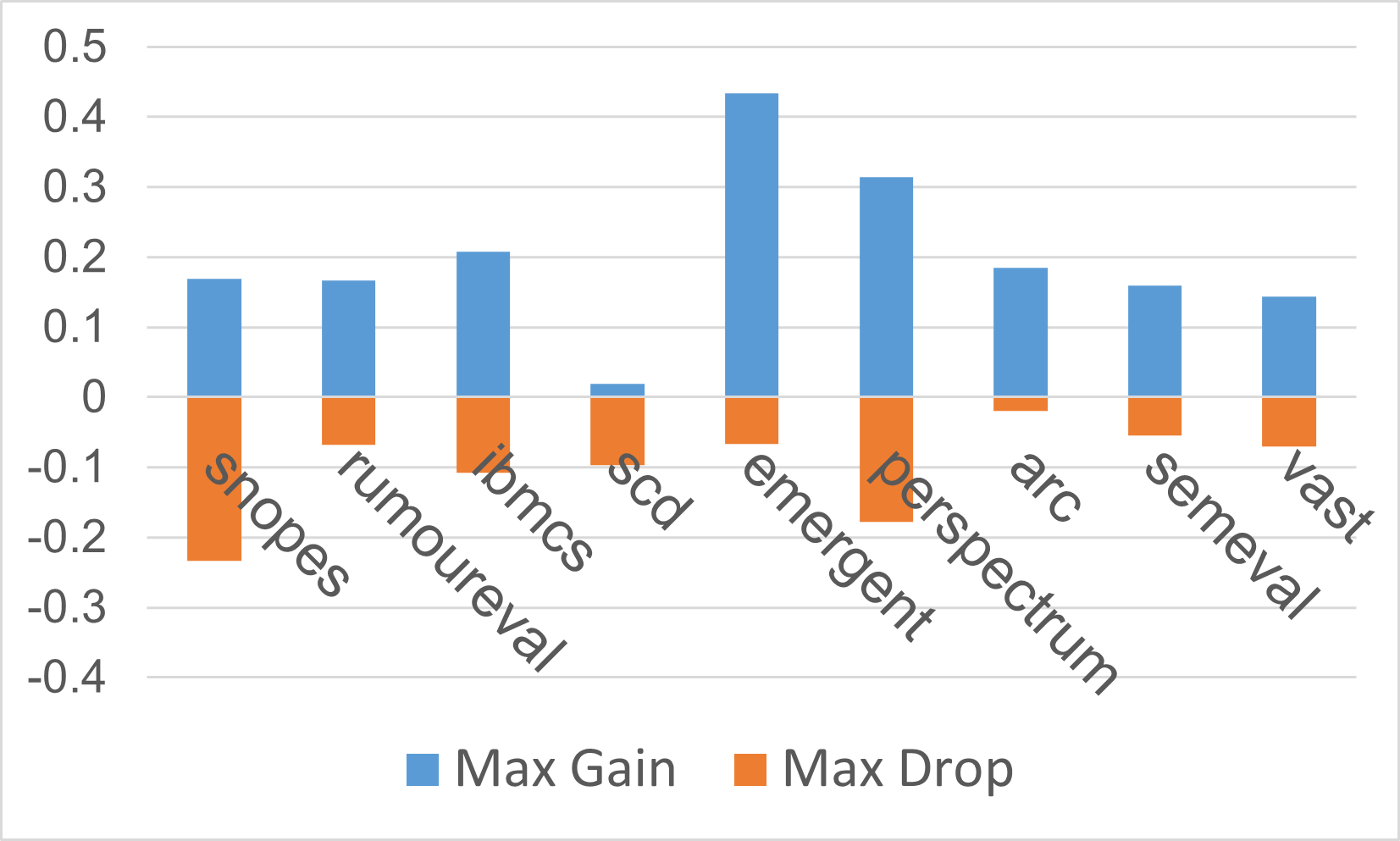}
\caption{Llama 3.1: top-k sub-optimal label sets}\label{subfig:llama31_topk-bad}
\end{subfigure}
\begin{subfigure}{0.35\textwidth}
\includegraphics[width=1\linewidth]{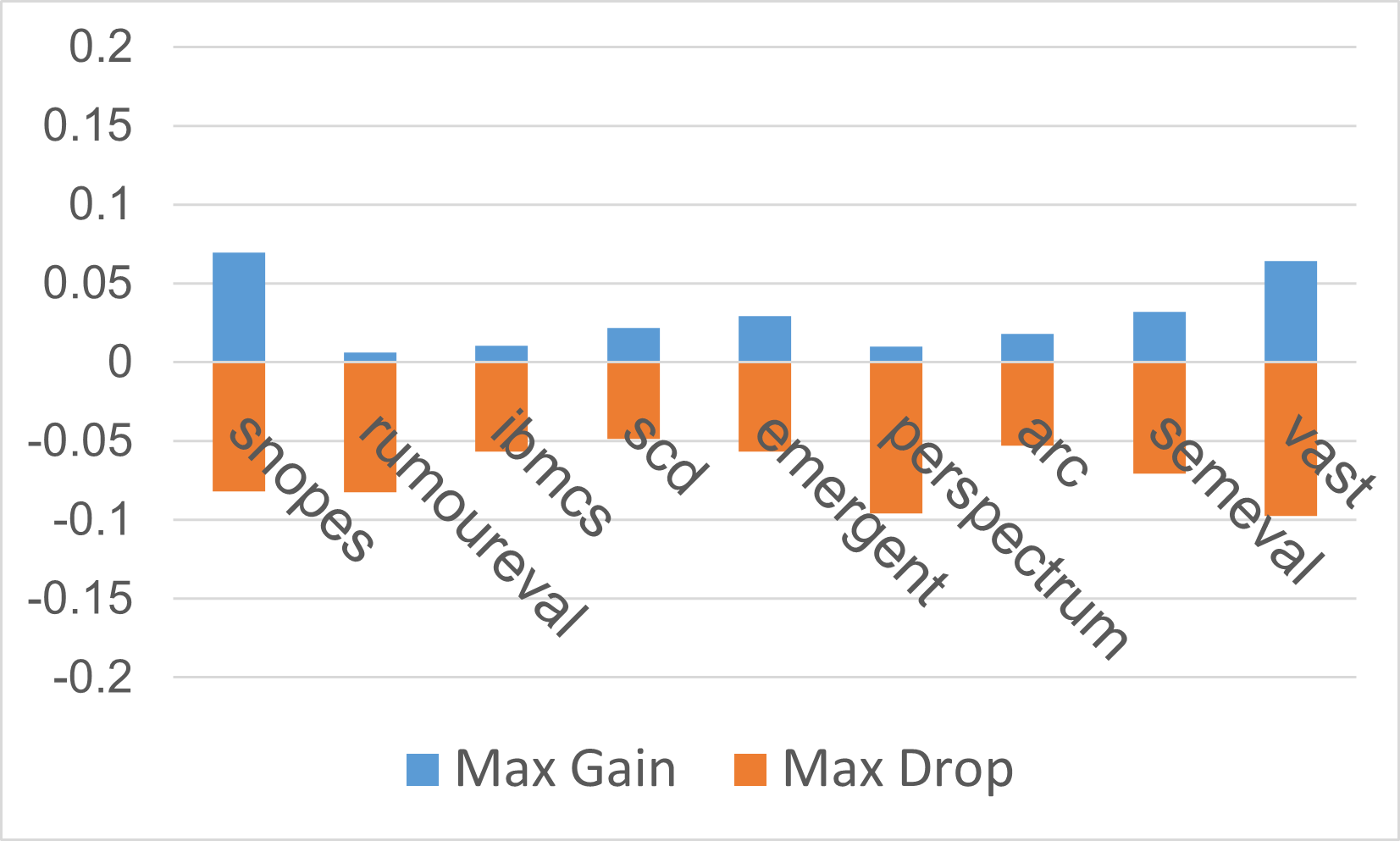}
\caption{Flan-T5-xl: top-k optimal label sets}\label{subfig:flan_topk-best}
\end{subfigure}
\begin{subfigure}{0.35\textwidth}
\includegraphics[width=1\linewidth]{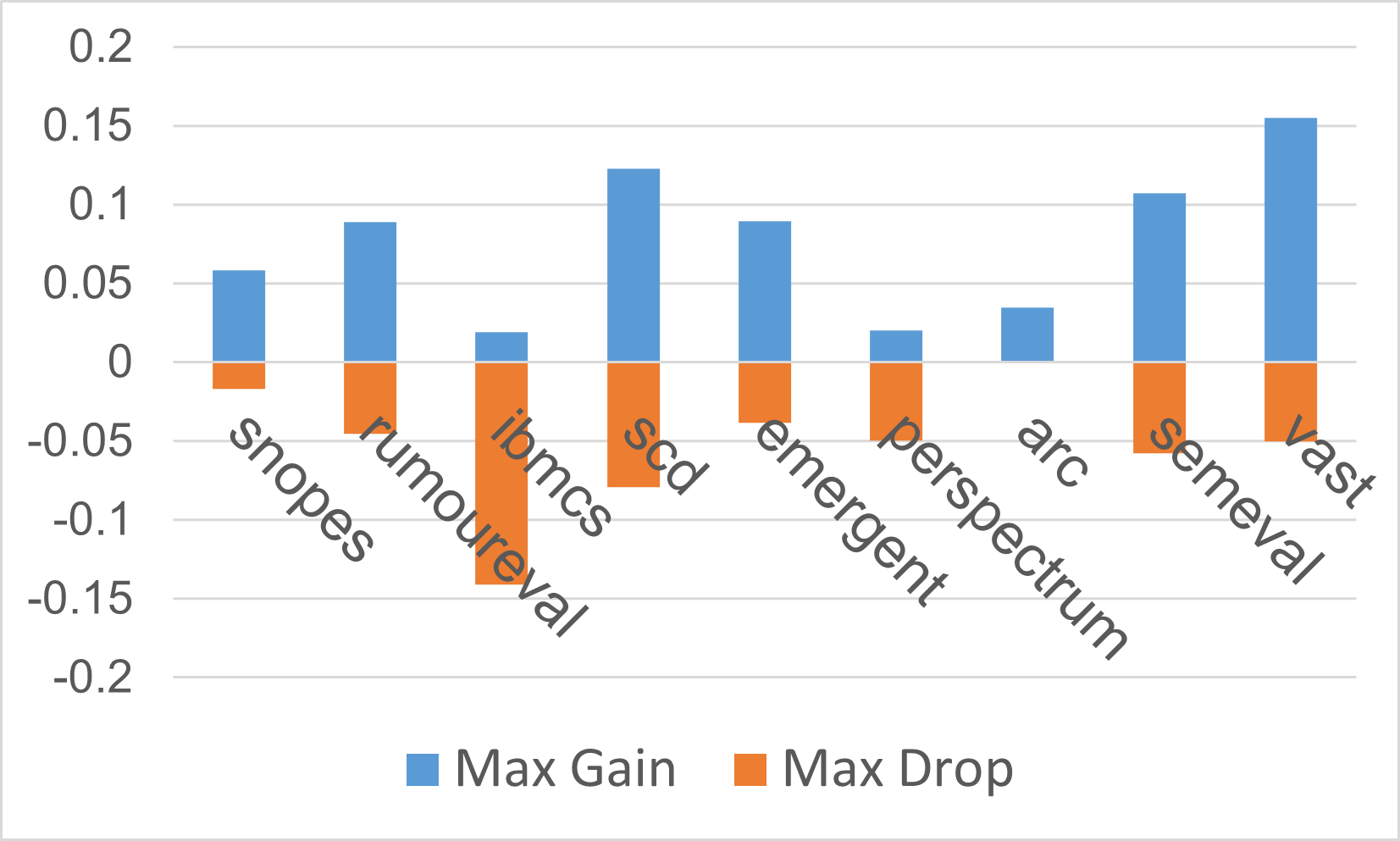}
\caption{Flan-T5-xl: top-k sub-optimal label sets}\label{subfig:flan_topk-bad}
\end{subfigure}

\caption{The maximum performance gain (positive value) and drop (negative value) on each dataset after re-ordering the label names for the top-k optimal and sub-optimal label sets with Llama3, Llama 3.1 and Flan-T5-xl.}
\label{fig:app label order}
\end{figure*}

The maximum performance gain and drop on each dataset after re-ordering the label names for the top-k optimal and poor label sets with Llama3, Llama 3.1 and Flan-T5-xl are in Figure \ref{fig:app label order}.

%\begin{figure}[t!]
%\centering

%\begin{subfigure}{0.23\textwidth}
%\includegraphics[width=1\linewidth]{image/llama_order_bestk.png}
%\caption{top-k optimal label sets}\label{subfig:topk-best}
%\end{subfigure}
%\begin{subfigure}{0.23\textwidth}
%\includegraphics[width=1\linewidth]{image/llama_order_worstk.png}
%\caption{top-k poor label sets}\label{subfig:topk-worst}
%\end{subfigure}
%\caption{The max performance gain (positive value) and drop (negative value) on each validation set after re-ordering the label names for the top-k optimal (a) and poor (b) label sets with Llama 3. Results on Flan-T5-xl and Llama 3.1 can be found in the Appendix.}
%\label{fig:label order llama3}
%\end{figure}

\section{Label Elaboration Results} \label{app:label ela}

We supplement the averaged absolute performance difference for each level of elaboration on Llama 3.1 and Flan-T5-xl in Table \ref{tab:apendix label length avg results}. 

%\begin{table}[h!]
%\centering
%\scalebox{0.6}{
%\begin{tabular}{l|cc|cc|cc}
%\hline
%\multirow{2}{4em}{\textbf{Dataset}} & \multicolumn{2}{c|}{$\mathbf{E_1}$} & \multicolumn{2}{c|}{$\mathbf{E_2}$} & \multicolumn{2}{c}{$\mathbf{E_3}$} \\
%\cline{2-7}
%& \textbf{Opt.} & \textbf{Sub-opt.} & \textbf{Opt.} & \textbf{Sub-opt.} & \textbf{Opt.} & \textbf{Sub-opt.}\\
%\hline
%perspectrum &  0.016 & 0.018 & 0.010 & 0.015 & 0.017 & 0.009\\
%ibmcs & 0.027 & 0.041 & 0.024 & 0.014 & 0.029 & 0.044\\
%snopes & 0.055 & 0.040 & 0.054 & 0.026 & 0.029 & 0.018\\
%emergent & 0.051 & 0.053 & 0.047 & 0.038 & 0.022 & 0.040\\
%rumoureval & 0.095 & 0.036 & 0.084 & 0.020 & 0.058 & 0.102\\
%arc & 0.017 & 0.024 & 0.015 & 0.021 & 0.035 & 0.049\\
%\hline
%\end{tabular}
%}
%\caption{The average absolute performance change after elaborating for \textit{optimal} (\textit{Opt.}) or \textit{poor} (\textit{Sub-opt.}) label sets with Llama 3 ($E_1$, $E_2$, $E_3$ see Figure \ref{fig:overview}). Results on Flan-T5-xl and Llama 3.1 can be found in the Appendix.}  
%\label{tab:elaboration}
%\end{table}

\begin{table}[h!]
\centering
\scalebox{0.63}{
\begin{tabular}{c|l|cc|cc|cc}
\hline
& Dataset & \multicolumn{2}{c|}{$E_1$} & \multicolumn{2}{c|}{$E_2$} & \multicolumn{2}{c}{$E_3$} \\
& & Opt. & Sub-opt. & Opt. & Sub-opt. & Opt. & Sub-opt.\\
\hline
%\hline
\multirow{6}{*}{\rotatebox[origin=c]{90}{Llama 3}} & perspectrum &  0.016 & 0.018 & 0.010 & 0.015 & 0.017 & 0.009\\
&ibmcs & 0.027 & 0.041 & 0.024 & 0.014 & 0.029 & 0.044\\
&snopes & 0.055 & 0.040 & 0.054 & 0.026 & 0.029 & 0.018\\
&emergent & 0.051 & 0.053 & 0.047 & 0.038 & 0.022 & 0.040\\
&rumoureval & 0.095 & 0.036 & 0.084 & 0.020 & 0.058 & 0.102\\
&arc & 0.017 & 0.024 & 0.015 & 0.021 & 0.035 & 0.049\\
\hline
\multirow{6}{*}{\rotatebox[origin=c]{90}{Llama 3.1}} & perspectrum &  0.027 & 0.020 & 0.048 & 0.025  & 0.037 & 0.022\\
& ibmcs & 0.033 & 0.018 & 0.054 & 0.031 & 0.057 & 0.067\\
& snopes & 0.015 & 0.021 & 0.032 & 0.020 & 0.034 & 0.033\\
& emergent & 0.048 & 0.077 & 0.048 & 0.107 & 0.034 & 0.133\\
& rumoureval & 0.041 & 0.040 & 0.037 & 0.032 & 0.039 & 0.035\\
& arc & 0.032 & 0.046 & 0.029 & 0.036 & 0.042 & 0.046\\
\hline
\multirow{6}{*}{\rotatebox[origin=c]{90}{Flan-T5-xl}} & perspectrum & 0.009 & 0.026 & 0.010 & 0.021 & 0.013 & 0.021\\
& ibmcs & 0.014 & 0.015 & 0.016 & 0.020  & 0.017 & 0.028\\
& snopes & 0.020 & 0.026 & 0.022 & 0.032 & 0.033 & 0.016\\
& emergent & 0.026 & 0.025 & 0.044 & 0.031 & 0.042 & 0.042\\
& rumoureval & 0.059 & 0.040 & 0.060 & 0.031 & 0.086 & 0.018\\
& arc & 0.034 & 0.042 & 0.025 & 0.038 & 0.031 & 0.041\\
\hline
\end{tabular}
}
\caption{The average absolute performance change after elaborating for \textit{optimal} or \textit{poor} single-word label sets with Llama 3.1 and Flan-t5-xl ($E_1$, $E_2$, $E_3$ see Figure 1 in main paper).}
\label{tab:apendix label length avg results}
\end{table}

\section{Prompt Sensitivity Analysis of \loads{}}\label{app:loads_sen}

To analyse the the prompt sensitivity of \loads{}, we test it on different prompt templates, select the label set through \loads{}, and then compare the performance with that on the label sets in the original dataset. 

Due to the computational resource constraints, we manually craft two prompts and test \loads{} with the four binary stance classification datasets on Llama 3. In the two prompts, we replace \textit{Given a [text1\_name] and a [text2\_name], detect the stance that the [text2\_name] has towards the [text1\_name]} (see Section 3.2 in main paper) with two different queries: 
\begin{enumerate}
    \item Prompt 1: \textit{What is the stance of [text2\_name] towards [text1\_name]?}
    \item Prompt 2: \textit{What stance does [text2\_name] take regarding [text1\_name]?}
\end{enumerate}

As shown in Table \ref{tab:prompt_sen}, although different prompts with the same label sets may result in performance changes as expected (compare with Table 5 in main paper), \loads{} is robust to different prompts used for the label selection. The performance gap between \loads{}-selected and original label sets tends to be similar across prompt templates.

\begin{table}[h!]
\centering
\scalebox{0.7}{
\begin{tabular}{ll|cc}
\hline
& \textbf{Dataset} & \textbf{LOADS} & \textbf{Original Label} \\
\hline
\multirow{4}{5em}{Prompt 1} & snopes & \underline{0.5926} & 0.4984\\
& ibmcs & \underline{0.8737} & 0.7303\\
& perspectrum & \underline{0.8925} & 0.8658\\
& scd & \underline{0.6895} & 0.6860\\
\hline
\multirow{4}{5em}{Prompt 2} & snopes & \underline{0.6191} & 0.5336\\
& ibmcs & \underline{0.8619} & 0.7523\\
& perspectrum & \underline{0.8921} & 0.8619\\
& scd & 0.6836 & \underline{0.6931}\\
\hline
\end{tabular}
}
\caption{Performance comparison on Llama 3 when using \loads{}-selected label sets (\textit{lowest kurtosis}) and using original label sets (\textit{original label}) with prompt 1 or prompt 2. The higher performance is underlined.}
\label{tab:prompt_sen}
\end{table}

\section{Perplexity Analysis} \label{app:ppl}

As discussed in Section 2 in main paper, Gonen et al.,\shortcite{gonen-etal-2023-demystifying} empirically show that zero-shot ICL performance is statistically negative correlated with the perplexity of the prompt with input. However, they did not take into account the label options in the prompt when calculating the perplexity. Therefore, we further investigate whether the perplexity is also correlated with the variance zero-shot ICL performance caused by different label names. 

Specifically, we use the prompt template in Section 3.2 in main paper, and calculate the perplexity of prompts with inputs and different label sets. Following Gonen et al.,\shortcite{gonen-etal-2023-demystifying}, for each label set, we average the perplexity over the dataset. And then we adopt spearman correlation test between the averaged perplexity scores and model performances. Due to the computational restriction, we experiment with all the binary datasets on Flan-T5-xl and Llama3-8b. Since Flan-T5 is an encoder-decoder model where perplexity has a loose definition, we treat the encoder input as an empty string when calculating perplexity.

The results in Table \ref{tab:app ppl} indicate that there is no statistically significant correlation between prompt perplexity and model performance if considering different label sets in the prompt.

\begin{table}[ht!]
\centering
\scalebox{0.7}{
\begin{tabular}{l|l|cccc}
\hline
&  & \textbf{perspectrum} & \textbf{ibmcs} & \textbf{snopes} & \textbf{scd}\\
\hline
\multirow{2}{5em}{Llama 3} & \textit{coefficient} & 0.0068 & 0.1641 & 0.0394 & 0.1698\\
 & \textit{p value} & 0.9707 & 0.3774 & 0.8303 & 0.3608\\
\hline
\multirow{2}{5em}{Flan-T5} & \textit{coefficient} & 0.0738 & 0.1733 & -0.0500 & -0.2273\\
& \textit{p value} & 0.6929 & 0.3511 & 0.7892 & 0.2187\\
\hline
\end{tabular}
}
\caption{Spearman correlation between model performance and prompt perplexity. P-values are all larger than 0.05, indicating no statistical significance.}
\label{tab:app ppl}
\end{table}

\section{Label Attention Key Similarity Analysis}  \label{app:labelkey}

In this section, we explore whether the closely related observation on few-shot ICL could be directly adopted to zero-shot ICL. Specifically, we focus on the study discussed in Section 2 in main paper, where Wang et al.,\shortcite{wang-etal-2023-label} suggest that the when the LLM is prompted by demonstration with examples in a few-shot ICL setting, the model is likely to confuse the label categories if their key vectors in the attention modules are similar to each other. 

Since this finding is easier to be tested on binary datasets, we experiment with the binary datasets on Llama 3 and Flan-T5-xl. We extract the key vectors in the attention module in each layer for each label name in the prompt. Then we calculate the cosine similarity between the vectors of two label names. Finally, we use spearman correlation test between similarity scores and model performances. As shown in Table \ref{tab:app labelkey}, we do not observe statistically significant correlation between model performance and label key vector similarities.

\begin{table}[ht!]
\centering
\scalebox{0.7}{
\begin{tabular}{l|l|cccc}
\hline
& & \textbf{perspectrum} & \textbf{ibmcs} & \textbf{snopes} & \textbf{scd}\\
\hline
\multirow{2}{5em}{Llama 3} & \textit{coefficient} & -0.0181 & -0.0051 & -0.2791 & -0.1696\\
 & \textit{p value} & 0.9244 & 0.9784 & 0.1283 & 0.3702\\
\hline
\multirow{2}{5em}{Flan-T5-xl} & \textit{coefficient} & -0.0595 & 0.0223 & -0.0209 & -0.0992\\
& \textit{p value} & 0.7503 & 0.9047 & 0.9108 & 0.5953\\
\hline
\end{tabular}
}
\caption{Spearman correlation between model performance and label's key vector similarity.}
\label{tab:app labelkey}
\end{table}

\section{Layer-wise Output Projections Analysis}

\begin{figure}[h!]
 \centering
 \includegraphics[width=0.43\textwidth]{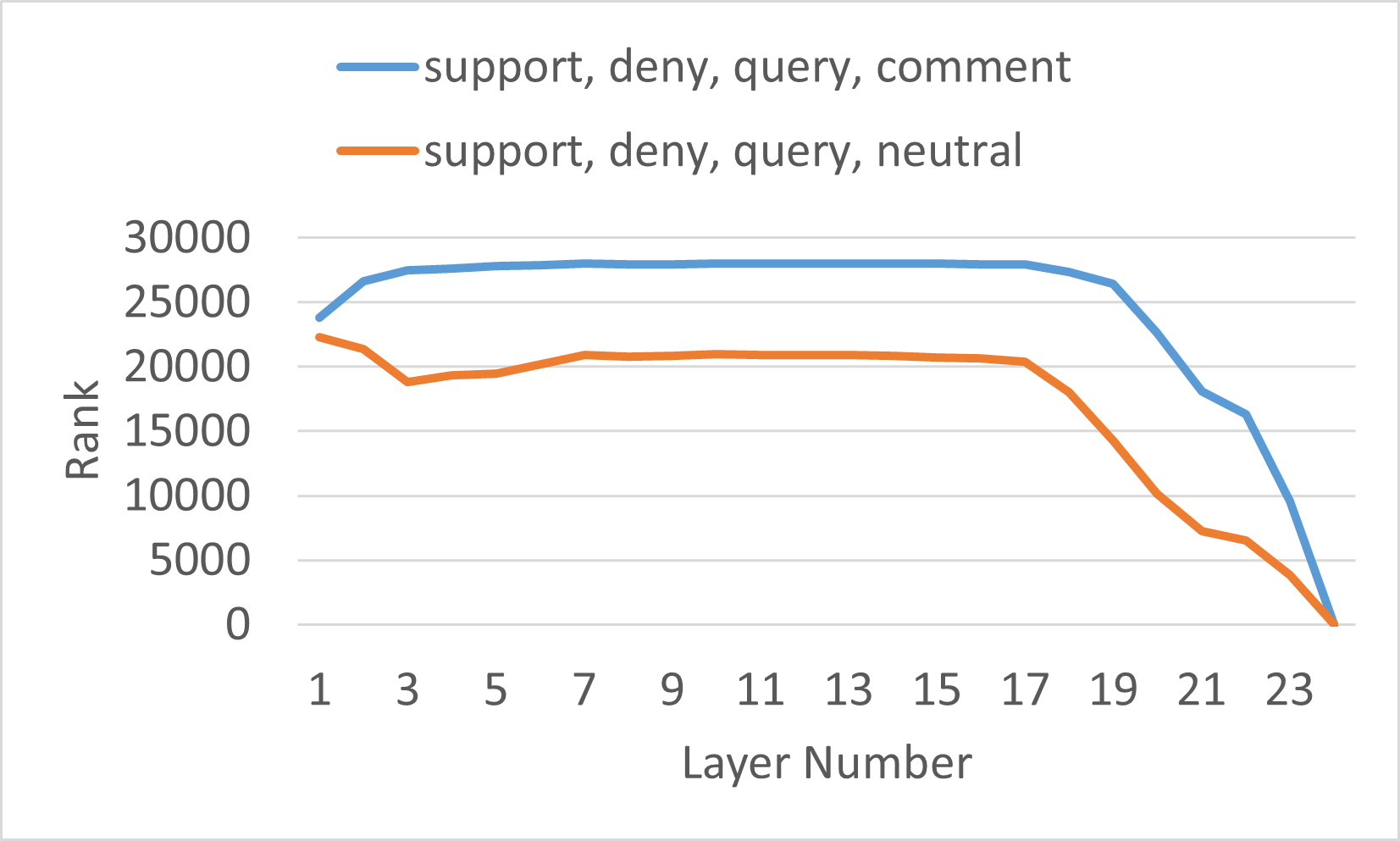}
 \caption{The rank of the final correctly predicted label (\textit{comment} or\textit{neutral}) when Flan-t5-xl is prompted with two different label sets for rumoureval dataset.} 
 \label{fig:app rank}
\end{figure}

We hypothesize that LLM may jump to the output prediction at last layers when a sub-optimal label set is used in the prompt. Therefore, we extract the hidden states from each decoder layer and project them on the model vocabulary, so that we obtain the ranked position of the final predicted label's token in each layer \cite{elhage2021mathematical,geva-etal-2022-transformer}. 

we observe that the hypothesis indeed holds in certain cases. We show an example on rumoureval dataset where we compare the averaged rank of the correctly predicted label \textit{comment/neutral} in each decoder layer when Flan-T5-xl is prompt to choose from \textit{support, deny, query, comment} or \textit{support, deny, query, neutral}. Label set \textit{support, deny, query, comment} performs worse than the set \textit{endorse, deny, query, neutral} on this dataset. As shown in Figure \ref{fig:app rank}, when using the relatively optimal label set \textit{endorse, deny, query, neutral}, the rank of the final predicted label tends to move closer to the top at an earlier stage.

\section{Human Translation Details}

To translate the English Twitter rumoureval test set into French and Portuguese, we recruit volunteer students from translation studies in Brazilian and French universities. The students are given gift vouchers (0.6 pounds per tweet). Consent has been obtained from the students and our study has received approval from the Ethics Committee of our university.

We instruct the students to translate the tweets accurately, and preserve the original meaning, context, and tone of the tweet. They are also encouraged to leave notes for their translations. The translations are finished on Google Sheets. 

\end{document}